\documentclass{article}

\usepackage[preprint]{neurips_2025}

\usepackage{natbib}
\setcitestyle{numbers,square}
\usepackage{multirow}
\usepackage{arydshln} 
\usepackage{subcaption}
\usepackage{makecell} 
\usepackage{placeins} 
\usepackage{graphicx}       
\usepackage{subcaption} 
\usepackage[bottom]{footmisc}
\usepackage{bbding}
\usepackage{soul}
\usepackage{hyperref} 
\usepackage{tablefootnote} 
\usepackage{caption} 
\usepackage{ulem}
\usepackage{cancel}
\usepackage{times}
\usepackage{booktabs} 
\usepackage{xcolor}
\usepackage{latexsym}
\usepackage{amssymb}
\usepackage[T1]{fontenc}
\usepackage{booktabs}
\usepackage{wrapfig}
\usepackage{tabularx}
\usepackage[utf8]{inputenc}
\usepackage{tcolorbox}
\usepackage{microtype}
\usepackage{amsmath}   
\usepackage{colortbl} 
\usepackage{inconsolata}
\usepackage{graphicx}
\usepackage{url}
\hypersetup{
    colorlinks=true,
    linkcolor=cyan,
    filecolor=blue,      
    urlcolor=cyan,
    citecolor=cyan,
}

\title{Dynamic Parametric Retrieval Augmented Generation for Test-time Knowledge Enhancement }

\newcommand{\github}{\raisebox{-1.5pt}{\includegraphics[height=1em]{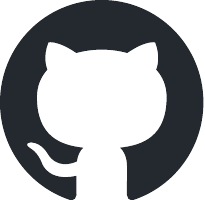}}}

\author{Yuqiao Tan$^{1,2}$, Shizhu He$^{1,2}$
\thanks{Corresponding author}, Huanxuan Liao$^{1,2}$, \textbf{Jun Zhao}$^{1,2}$, \textbf{Kang Liu}$^{1,2}$  \\
    $^1$ The Key Laboratory of Cognition and Decision Intelligence for Complex Systems, \\
    Institute of Automation, Chinese Academy of Sciences, Beijing, China \\
    $^2$ School of Artificial Intelligence, University of Chinese Academy of Sciences, Beijing, China \\
  {\{tanyuqiao2025, liaohuanxuan2023\}@ia.ac.cn} {\{shizhu.he, jzhao, kliu\}@nlpr.ia.ac.cn} \\
 {\small{\github{} \url{https://github.com/Trae1ounG/DyPRAG}}}
  }

\begin{document}

\maketitle

\begin{abstract}
\label{sec:abs}
Retrieval-augmented generation (RAG) enhances large language models (LLMs) by retrieving relevant documents from external sources and incorporating them into the context. While it improves reliability by providing factual texts, it significantly increases inference costs as context length grows and introduces challenging issue of RAG hallucination, primarily caused by the lack of corresponding parametric knowledge in LLMs. An efficient solution is to enhance the knowledge of LLMs at test-time.
Parametric RAG (PRAG) addresses this by embedding document into LLMs parameters to perform test-time knowledge enhancement, effectively reducing inference costs through offline training. However, its high training and storage costs, along with limited generalization ability, significantly restrict its practical adoption. 
To address these challenges, we propose \textbf{Dynamic Parametric RAG (DyPRAG)}, a novel framework that leverages a lightweight parameter translator model to efficiently convert documents into parametric knowledge.
DyPRAG not only reduces inference, training, and storage costs but also dynamically generates parametric knowledge, seamlessly enhancing the knowledge of LLMs and resolving knowledge conflicts in a plug-and-play manner at test-time. Extensive experiments on multiple datasets demonstrate the effectiveness and generalization capabilities of DyPRAG, offering a powerful and practical RAG paradigm which enables superior knowledge fusion and mitigates RAG hallucination in real-world applications.
\end{abstract}

\section{Introduction}
\label{sec:intro}

Knowledge-intensive tasks, such as question answering (QA), require leveraging extensive world and domain knowledge \cite{tqa, nq}. While Large Language Models (LLMs) have demonstrated impressive capabilities across various tasks and fields \cite{pre}, they struggle to handle knowledge-intensive tasks independently due to limitations in accessing external knowledge and a tendency to generate hallucinated information \cite{frisoni2024generate}.

\begin{wrapfigure}[16]{r}{0.5\textwidth}
\centering
\includegraphics[width=0.5\textwidth]{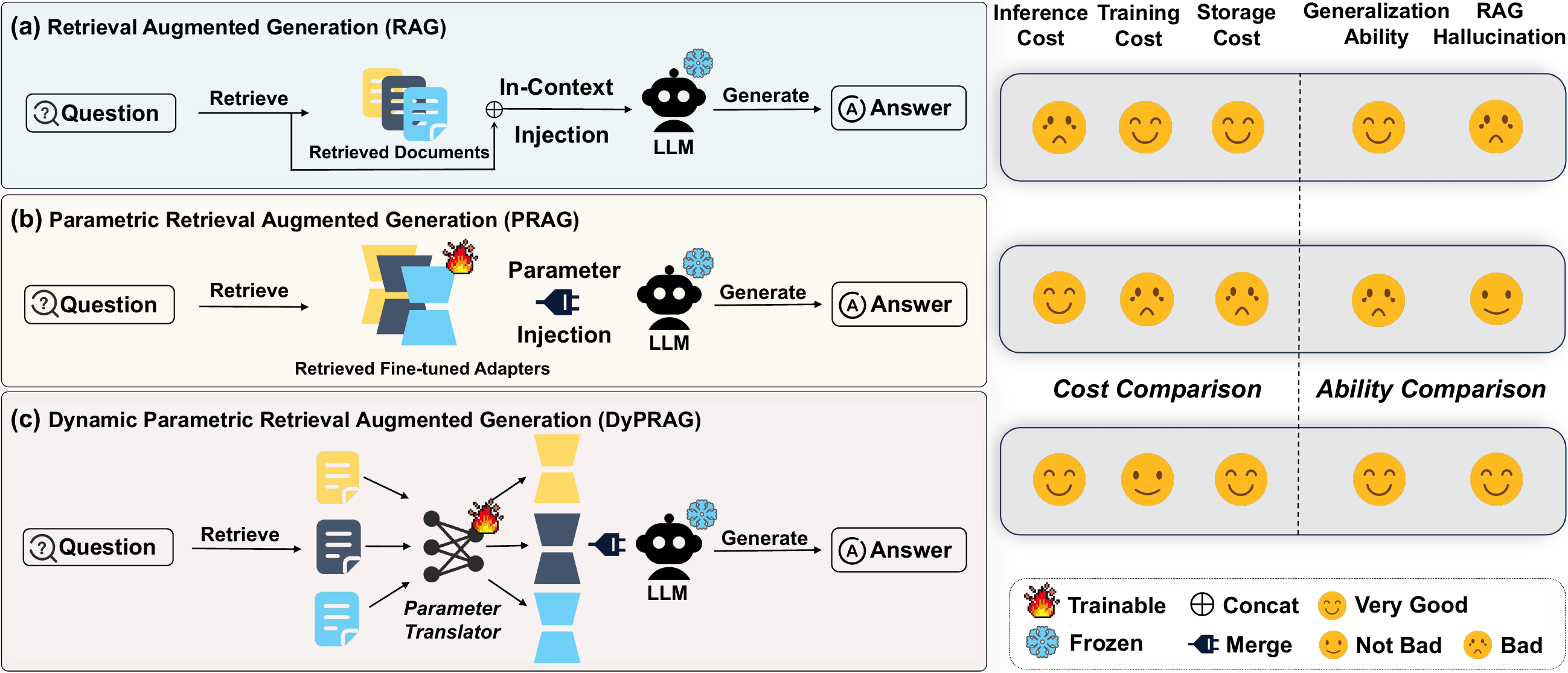}
\caption{
Compared to RAG and PRAG, the proposed DyPRAG offers multiple advantages, including lower inference, training and storage cost, strong generalization ability, and mitigation of RAG hallucination.
}
\label{fig:intro}
% \vspace{-1cm}
\end{wrapfigure}

To mitigate incorrect internal knowledge and improve performance in QA, researchers have developed knowledge-augmented methods for LLMs. One of the most used methods called Retrieval-Augmented Generation (RAG) \cite{REALM}, which retrieves relevant documents from external sources (e.g., Wikipedia) and integrates both the retrieved documents and the query into the model’s input \cite{fid} (as shown in Figure~\ref{fig:intro} (a)).
Standard RAG methods~\cite{asaiself,jiang2022retrieval,su2024mitigating} share a common approach: injecting external knowledge into LLMs by appending retrieved documents to the input context, referred to as \textbf{in-context injection}. This technique harnesses the reasoning capabilities of LLMs to address knowledge gaps and reduce misinformation~\cite{brown2020languagemodelsfewshotlearners}. However, it introduces several challenges. As the number and length of retrieved documents increase, the extended context leads to significantly higher inference costs. Additionally, when the internal knowledge of LLMs conflicts with the external context, a phenomenon known as RAG hallucination (or knowledge conflict) arises, causing LLMs to produce incorrect answers even when accurate documents are provided~\cite{zhang2024evaluatingexternalparametricknowledge,sun2025redeepdetectinghallucinationretrievalaugmented}.

% \begin{wrapfigure}[16]{r}{0.6\textwidth}
% \includegraphics[width=0.6\textwidth]{./fig/dyprag_Intro.pdf}
% \caption{
% Compared to RAG and PRAG, the proposed DyPRAG offers multiple advantages, including lower inference, training and storage cost, strong generalization ability, and mitigation of RAG hallucination.
% }
% \label{fig:intro}
% \vspace{-1cm}
% \end{wrapfigure}

Parametric RAG (PRAG) employs another way of injecting knowledge which integrates external knowledge directly into the parameters of LLMs, known as \textbf{parameter injection} (in Figure~\ref{fig:intro} (b)). PRAG first rewrites retrieved documents and transforms them into QA pairs, facilitating knowledge memorization and manipulation~\cite{allenzhu2024physicslanguagemodels31, allenzhu2024physicslanguagemodels32}. 
The workflow of PRAG is divided into two stages. During the offline phase, these augmented documents are fine-tuned with LoRA~\cite{hu2021loralowrankadaptationlarge}, encoding contextual knowledge directly into parameters. In the online phase, retrieved documents are replaced with loadable parameters.
% , reducing context length while embedding external knowledge into the internal representation of LLMs for test-time knowledge enhancement.
However, this approach has significant trade-offs. Augmenting, training, and storing parameters for each retrieved document incurs high computational and storage costs, coupled with non-generalization, severely limiting scalability in real-world applications (e.g., systems in domains requiring frequent knowledge updates or multi open-domain QA tasks). This presents a critical challenge for PRAG: \textit{\textbf{how can we minimize the overhead of converting documents into parametric knowledge without compromising performance?}}

% \begin{wrapfigure}[16]{r}{0.6\textwidth}
% \includegraphics[width=0.6\textwidth]{./fig/dyprag_Intro.pdf}
% \caption{
% Compared to RAG and PRAG, the proposed DyPRAG offers multiple advantages, including lower inference, training and storage cost, strong generalization ability, and mitigation of RAG hallucination.
% }
% \label{fig:intro}
% \vspace{-1cm}
% \end{wrapfigure}

In this work, we propose \textbf{Dynamic Parametric RAG (DyPRAG)},  a lightweight framework that converts documents into parameters directly, making DyPRAG seamlessly enhance the knowledge of LLMs in a plug-and-play manner at test-time (as shown in Figure~\ref{fig:intro} (c)). 
Rethinking of PRAG, its intrinsic goal is to obtain an underlying mapping function $\mathcal{F}$ which transforms external documents into parameters through augmentation and training for each document. 
% We believe better wit than wealth—\textit{the ability to convert documents into parameters holds greater significance than arduously storing a large collection of parametric representations}.
Building on this insight, we aim to establish a more generalized mapping function by directly modeling $\mathcal{F}$ with a small hypernetwork $\mathcal{F}^\prime_\phi$, referred to as parameter translator.
With well-trained $\mathcal{F}^\prime_\phi$, DyPRAG eliminates the need for the cumbersome training and storage process required by PRAG. After a detailed analysis of computation and storage overhead, our method significantly reduces the high inference costs of traditional RAG while eliminating the rigid training and storage costs of PRAG.

Through extensive experiments, we derive two key findings: \textbf{1) DyPRAG enhances the test-time parametric knowledge of LLMs effectively.}  During evaluation,  DyPRAG outperforms standard RAG across different scales of LLMs, demonstrating its ability to enhance the internal knowledge of LLMs. 
Furthermore, although PRAG learns $\mathcal{F}$ by training separately for each document,
DyPRAG achieves comparable or even better performance in various scenarios with significantly lower costs.
\textbf{2) Combining test-time generated parametric knowledge with contextual knowledge leads to superior knowledge fusion.} Following ~\cite{su2025parametricretrievalaugmentedgeneration}, we further investigate the combination of in-context injection and parameter injection, referred to as DyPRAG-Combine. In both independent identically distributed and out-of-distribution settings, DyPRAG-Combine achieves  the best results, showing strong generalization ability. Additionally, we find that DyPRAG-Combine effectively relieves RAG hallucination by first injecting context-related parameters to reduce the knowledge conflicts.  Based on the experimental results on RAGTruth~\cite{niu2023ragtruth} benchmark, we observe that DyPRAG-Combine enables LLMs to internalize previous unseen knowledge. 
This suggests that integrating parametric and contextual knowledge using DyPRAG could be a powerful approach for building a trustworthy RAG system in real-world applications. We summarize our contributions as follows:

\begin{itemize}
    \item We propose Dynamic Parametric RAG (DyPRAG), a lightweight framework to efficiently convert documents into parameters in a plug-and-play manner with minimal cost.
    \item We introduce a powerful and practical RAG paradigm that effectively combines contextual knowledge with test-time generated parametric knowledge, enabling superior knowledge fusion.
    \item Experimental results show that DyPRAG excels in generalization, efficiently injects parameters, and seamlessly incorporates contextual knowledge, enhancing performance while reducing RAG hallucination.
\end{itemize}

\section{Related Work}
\subsection{Retrieval Augmented Generation}
Large language models (LLMs) have demonstrated remarkable performance across diverse applications. However, their inherent knowledge often falls short in handling knowledge-intensive tasks, highlighting the need for external knowledge integration to ensure robust performance in such contexts.
A prominent approach to bridging this gap is Retrieval-Augmented Generation (RAG), which enhances LLMs by incorporating relevant external knowledge sources~\cite{borgeaud2022improving, wang2024knowledge, wang2024lekube, guu2020retrieval}. The retrieved documents are appended to the LLM's input context, enabling it to leverage knowledge beyond its training data~\cite{lewis2020retrieval}. We refer to this as the in-context  injection paradigm. However, this approach leads to high inference costs and can exceed context length limits as the number and length of retrieved documents increase~\cite{xiong2023effectivelongcontextscalingfoundation}.
To address this issue, recent study introduces Parametric RAG (PRAG)~\cite{su2025parametricretrievalaugmentedgeneration}, a  paradigm that fine-tunes the model on augmented documents, encoding useful information into its parameters. While PRAG mitigates the inference cost, it introduces additional training and storage costs due to the need to obtain and store LoRA parameters.
Our proposed method significantly reduces the high costs associated with standard RAG and PRAG while achieving superior generalization. By combining contextual knowledge with test-time generated parametric knowledge through DyPRAG, it enables LLMs to better manipulate knowledge, effectively mitigating RAG hallucination. This makes it a more practical and robust solution for real-world applications.
\subsection{Context Compression}
Context compression is widely adopted to improve the efficiency of LLMs in processing contextual knowledge. Recent studies \cite{learning, ge2024incontextautoencodercontextcompression, dai2024muap} propose condensing long contexts into soft prompts, allowing LLMs to utilize information more effectively. Meanwhile, other works 
~\cite{mao2024liftimprovinglongcontext, wang2024greatertextcomesgreater} focus on transforming context chunk into LoRA modules to improve the understanding ability of extended contexts while xRAG
\cite{cheng2024xragextremecontextcompression} integrates context compression by mapping documents into a compact token representation. Similarly, AAG~\cite{liao2024awakeningaugmentedgenerationlearning} draws inspiration from human cognition, retrieving and recalling relevant knowledge to compensate for knowledge gaps. This approach activates relevant information within LLMs without relying on external resources.
Our proposed method dynamically converts retrieved documents into  parameters, effectively unifying internal and external knowledge while mitigating RAG hallucination issue. Furthermore, it functions as a plug-and-play solution to enhance the efficiency and adaptability of standard RAG systems by test-time parametric knowledge enhancement.
\section{Methodology}
\begin{figure*}[t]
\centerline{\includegraphics[width=\textwidth]{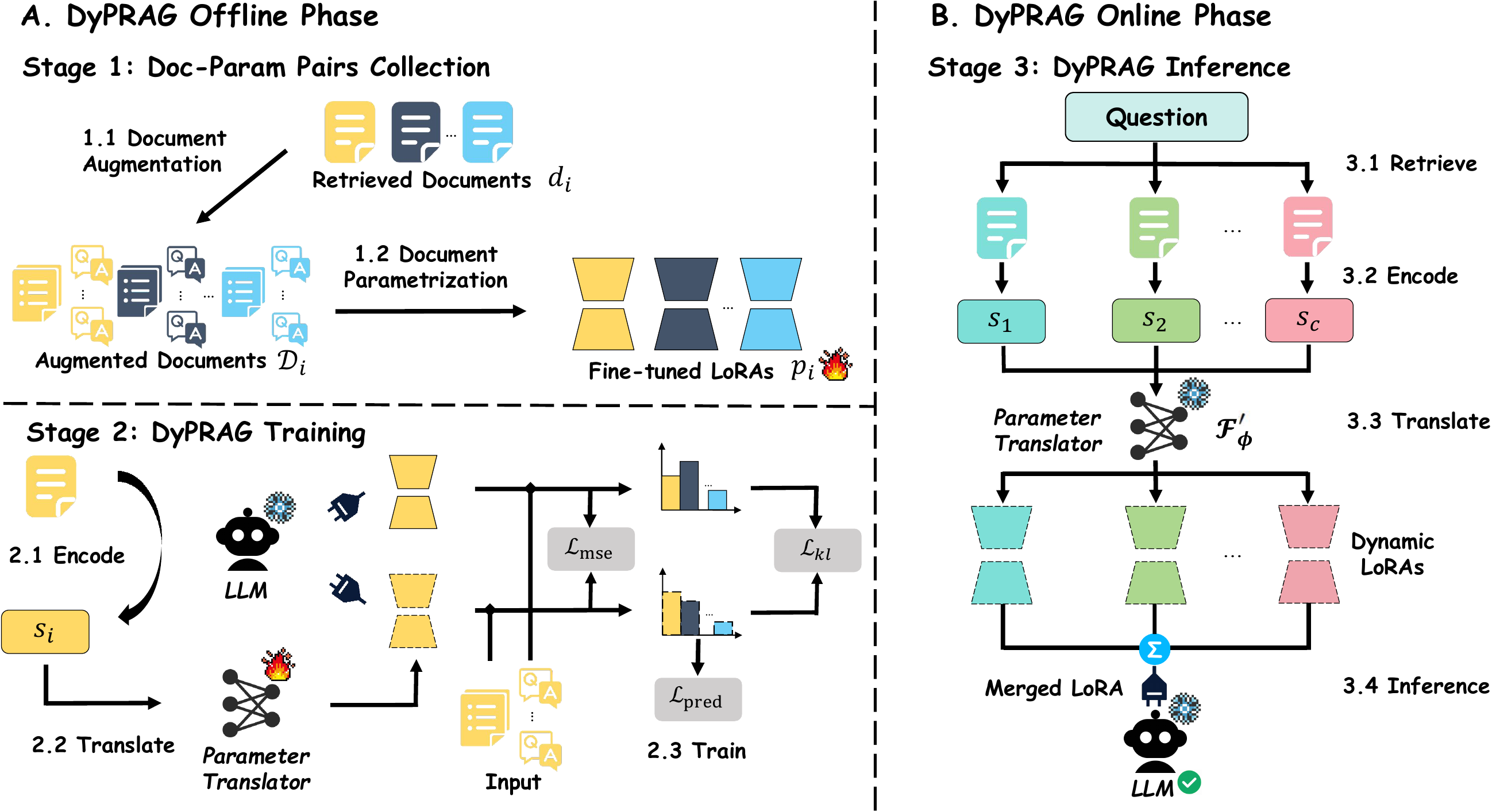}}
% \vspace{-0.45cm}
\caption{
An illustration of the DyPRAG method. In the offline phase, \textbf{Stage 1} follows the same parameterization process as PRAG to collect Doc-Param pairs. In \textbf{Stage 2}, we train the parameter translator $\mathcal{F}^\prime_\phi$ to learn the mapping function from documents to parameters. During the online \textbf{Stage 3}, the trained $\mathcal{F}^\prime_\phi$ dynamically generates LoRA modules to enhance LLMs knowledge at test-time.
}
\label{fig:method}
\end{figure*}

\label{sec:method}
In this section, we introduce the Dynamic Parametric RAG framework, as shown in Figure~\ref{fig:method}. We first formulate the problem and review the previous PRAG framework. Specifically, we revisit the offline document parameterization process, which transforms documents into parametric representations through \textbf{Document Augmentation} and \textbf{Document Parameterizing}, following~\cite{su2025parametricretrievalaugmentedgeneration}.
Subsequently, we present our \textbf{Parameter Translation} process, which learns the underlying function to map document embeddings into adapter parameters (e.g., LoRA~\cite{hu2021loralowrankadaptationlarge}). Once the translator is well optimized, retrieved documents can be directly converted into parametric representations. These representations can then be efficiently integrated into the LLMs, enhancing response quality while reducing inference, training, and storage costs at test-time.

\subsection{Preliminary of Parametric RAG}
\label{sec:prag}
This subsection introduces the problem formulation of the RAG task and outlines the static Parametric RAG pipeline proposed in~\cite{su2025parametricretrievalaugmentedgeneration}. 
\paragraph{\noindent\textbf{Standard RAG.}}
Let $\mathcal{M}$ denote a large language model (LLM) with base parameters $\Theta$. Given a user query $q$, the task is to generate an accurate response informed by an external corpus $\mathcal{C}$, expressed as $\mathcal{C} = \{d_1, d_2, \ldots, d_N\}$. Each element $d_i$ , referred to as a document, represents a text chunk retrieved from Wikipedia in our experiments~\cite{fid}. To achieve this, a retrieval module $\mathcal{R}$ is employed to compute relevance scores between $q$ and the documents in $\mathcal{C}$. The traditional RAG approach selects the top-$c$ documents with the highest similarity scores and concatenates them with the user query to form the input context. Using this augmented input, $\mathcal{M}$ generates the response by leveraging both the query and the retrieved knowledge. This procedure, referred to as \textbf{In-context Injection}, significantly increases inference costs as the context length and the number of selected documents grow.

\paragraph{\noindent\textbf{Parametric RAG.}}
In contrast, Parametric RAG (PRAG) integrates documents directly into the parameters of $\mathcal{M}$ to reduce the cost associated with long contexts. Each document $d_i \in \mathcal{C}$ is transformed offline into a parametric representation $p_i = \mathcal{F}(d_i)$, where $\mathcal{F}$ is a underlying mapping function that converts each document $d_i$ into its corresponding parametric representation $p_i$. To achieve a more effective mapping, PRAG employs \textbf{Document Augmentation}, inspired by~\cite{allenzhu2024physicslanguagemodels31}, to help the model memorize and manipulate the information contained in the document $d_i$. Specifically, PRAG uses $\mathcal{M}$ to rewrite $d_i$ into multiple variations, resulting in $\{d^1_i, d^2_i, \dots, d^n_i\}$. Additionally, for each original document $d_i$, PRAG prompts $\mathcal{M}$ to generate $m$ question-answer (QA) pairs: $\{(q^1_i, a^1_i), (q^2_i, a^2_i), \dots, (q^m_i, a^m_i)\}$, where $n$ and $m$ are controllable hyperparameters. 
This augmented set of documents preserves the factual content of the original document while incorporating diverse linguistic variations, expressed as:
\begin{equation}
\mathcal{D}_i = \big\{({d_i}^k, {q_i}^j, {a_i}^j) \,\big|\, k \in [1, n],\, j \in [1, m] \big\},    
\end{equation}
where each $(d^k_i, q^j_i, a^j_i)$ triple is then concatenated to a training sample $x=[d^k_i \oplus q^j_i \oplus a^j_i]$. For \textbf{Document Parameterizing}, PRAG utilizes LoRA~\cite{hu2021loralowrankadaptationlarge} approach to encode parametric knowledge for each $\mathcal{D}_i$ where the overall goal is to optimize:
\begin{align}
    \label{eq:lm_training}
    \min\limits_{{\Delta \Theta}} \sum\limits_{{\substack{({d_i}^k, {q_i}^j, {a_i}^j) \in {D}_i}}} 
    \sum\limits_{{t=1}}^{T} 
    -\log \; P_{{\substack{\Theta + \Delta \Theta}}}\bigl(x_t \,\big|\, x_{<t}\bigr),
\end{align}
where $\Delta\Theta$ is the trainable low-rank matrix and only apply to feed-forward network (FFN), we refer this paradigm to as \textbf{Parameter Injection}. Notably, although this method eliminates the use of documents as context, it further introduces significant training cost and storage cost, which will be analyzed in later section.

\subsection{Dynamic Parametric RAG}
In this section, we describe the detailed process of our newly proposed Dynamic Parametric RAG (DyPRAG) paradigm. To summarize the previous PRAG, its primary goal is to obtain a well-performing mapping function $\mathcal{F}$ through augmentation and training for each document $d_i$. However, this intricate process is impractical for real-world scenarios as it requires repeating these steps and consuming significant computing resources and time when faced with new documents or queries. We argue that the key to optimization lies in addressing the question: \textbf{\textit{How to obtain a generalized mapping function $\mathcal{F}$?}} To solve this, we propose a three-stage dynamic framework designed to enable parameter injection in an effective and efficient manner.

\paragraph{\noindent\textbf{Doc-Param Pairs Collection.}} 
To derive the general mapping function $\mathcal{F}$, we start by collecting a set of document-parameter (Doc-Param) pairs using the method described in Sec.~\ref{sec:prag}. For each document $d_i$, we collect its corresponding parametric representation $p_i$, forming the alignment set $\mathcal{K} = \{(d_1, p_1), (d_2, p_2), \dots, (d_N, p_N)\}$.

\paragraph{\noindent\textbf{Dynamic Parametric RAG Training.}}
\label{sec:train}
After obtaining the alignment set $\mathcal{K}$, we utilize the original LLM $\mathcal{M}$ to encode textual documents into embeddings. For a given document $d_i$, we extract the last hidden state $s_i \in \mathbb{R}^h$ at the final token position before transforming it into the vocabulary space, where $h$ represents the hidden dimension of $\mathcal{M}$. 
To model the implicit relationship between documents and parameters, we design a simple hypernetwork called \textbf{Parameter Translator} $\mathcal{F}^\prime_\phi$ to translate $s_i$ into parametric representation. This hypernetwork consists of several linear layers parameterized by a base parameter $\phi$. As an example, consider the up-project module in FFN. The standard LoRA process as follows:
\begin{equation}
    W^\prime=W+\Delta W=W+BA
\end{equation}
where $W\in \mathbb{R}^{h\times k}$, $B\in \mathbb{R}^{h\times r}$ and $A \in \mathbb{R}^{r \times
 k}$. $k$ represents the intermediate dimension of FFN and $r$ is the controllable rank of LoRA matrices. During training phase, $\mathcal{F}^\prime_\phi$ performs separately on $B$ and $A$. Formally:
 \begin{equation}
     B^l =  \text{Reshape}(W^{l,B}_{up}\text{Relu}(W^{l,B}_{down}(s_i \oplus idx^l)))
     \label{eq:parameter_translator}
 \end{equation}
 where $W^{l,B}_{down} \in \mathbb{R}^{p\times (h+1)}$ and $W^{l,B}_{up} \in \mathbb{R}^{hr\times p}$. Here, $p$ represents the tunable intermediate dimension of the MLP module in $\mathcal{F}^\prime_\phi$, and $\text{Reshape}(\cdot)$ reshapes the output vector into the shape of $B$. 
This process is applied at each layer $l$, so we concatenate the layer index with $s_i$. We provide the visualization of this workflow in Figure~\ref{fig:parameter_translator}. A similar procedure is followed for matrices $A$ and in other modules of FFN. The parametric representation generated by $\mathcal{F}_\phi^\prime$ is denoted as $p^\prime_i$. Our goal is for it to perform as effectively as $p_i$.

 To align with PRAG~\cite{su2025parametricretrievalaugmentedgeneration}, we utilize the augmented dataset $\mathcal{D}_i$ and the same objective function as presented in Eq.~\ref{eq:lm_training} to optimize $\mathcal{F}^\prime_{\phi}$ which corresponds to $\mathcal{L}_{\text{pred}}$. Additionally, for the target LoRA adapter $p_i$, we employ $\mathcal{L}_{\text{mse}}$ to compute the difference between the generated parameters and the target parameters. The Kullback-Leibler divergence, denoted as $\mathcal{L}_{\text{kl}}$, quantifies the discrepancy in word probability distributions between the two models, with the model using $p_i$ serving as the target distribution to be imitated. The overall formulation is given by:
 \begin{equation}
     \mathcal{L}_{\text{mse}}=\text{MSE}(p_i, \mathcal{F}^\prime_\phi(d_i))
 \end{equation}
 \begin{equation}
     \mathcal{L}_{\text{kl}}=\text{KL}(P_{\Theta+p_i}(x\mid D_i), P_{\Theta+\mathcal{F}^\prime_{\phi}(d_i)}(x \mid D_i))
\end{equation}
\begin{equation}
    \mathcal{L}_{\text{align}}=\mathcal{L}_{\text{pred}}+\lambda_1 \mathcal{L}_{\text{mse}}+\lambda_2 \mathcal{L}_{\text{kl}}
\end{equation}
where we calculate the overall alignment loss for each document $d_i$, $\lambda_1$ and $\lambda_2$ are tunable hyper-parameter which set to 100 and 0.01 separately to make loss range similar.

\paragraph{\noindent\textbf{Dynamic Parametric RAG Inference.}}
During the inference stage, once a well-trained parameter translator $\mathcal{F}^\prime_{\phi}$ is obtained, we can efficiently perform parameter injection while significantly reducing inference costs. For a test query $q^{t}$, we rerun the retrieval process using the retrieval module $\mathcal{R}$ to select the most relevant documents. For each selected document $d^t_i$, we derive its embedding $s^t_i$ and input it into $\mathcal{F}^\prime_{\phi}$ to obtain the dynamic LoRA adapter $p^{t,\prime}_i$, which encodes the relevant information from the document in parameter modality~\cite{cheng2024xragextremecontextcompression}. We then merge this as the LoRA parameter for inference, resulting in low inference costs without requiring a concatenated context.

\subsection{Computation and Storage Cost Analysis}
We present an initial pilot analysis and a broad evaluation of computation and storage costs across three baseline methods. More detailed analysis of time complexity is provided in Appendix~\ref{app:detail_compare}.
\label{sec:cost_ana}
\paragraph{\noindent\textbf{Computation Cost.}} 
The computation cost in  RAG is primarily the in-context inference cost, whereas PRAG introduces additional training and inference costs due to augmentation and offline parametrization. Suppose the average token count of document $d$ is $|d|$. As noted in~\cite{su2025parametricretrievalaugmentedgeneration}, the augmentation process typically generates about $2|d|$ tokens, leading to an augmentation cost of $3|d|$. 
When training the target LoRA, a forward pass over $3|d|$ tokens and a backward pass over $6|d|$ tokens (typically twice the forward pass cost) result in a total training cost of $9|d|$. 
Although these tasks can be performed offline, it still requires a long time and do not generalize to new questions with newly retrieved documents. 
In contrast, our DyPRAG approach offers a more practical solution by requiring only $N$ Doc-Param pairs while even a small $N$ can achieve powerful performance, significantly reducing costs for augmentation and training. 
The cost of MLP-based $\mathcal{F}^\prime_{\phi}$ 
is negligible compared to Transformer-based LLMs \cite{vaswani2023attentionneed}. 

The primary advantage of PRAG is the reduction of inference cost. Let $|q|$ denote the length of the question, $c$ represent the number of retrieved documents. The inference context of PRAG and DyPRAG is $|q|$, whereas in-context injection RAG requires $c|d|+|q|$. The parameterization process significantly reduces the inference cost, especially when $|d|$ and $c$ grow larger. Notably, the inference cost is also closely tied to the length of model response. DyPRAG demonstrates an improved ability to internalize knowledge, resulting in shorter responses that effectively reduce  costs (in Figure~\ref{fig:qwen_res_length}).

\paragraph{\noindent\textbf{Storage Cost.}}
One of the main shortcomings of PRAG is the storage cost associated with $p_i$. Let $r$ denote the LoRA rank, $L$ the number of Transformer layers, $h$ the hidden size, and $k$ the intermediate size of the FFN. The number of parameters in the parametric representation of a document is  $6Lr(h+k)$. For instance, in the Qwen2.5-1.5B model (which has 28 layers, a hidden dimension of 1536, and an intermediate size of 8960), setting $r$ to 2 results in approximately 3.53M parameters, storing 6.73MB in 16-bit precision for each parametric representation. In our following experiments, we need to store 18.66GB offline parameters for Qwen2.5-1.5B, presenting a significant cost.

In contrast, our DyPRAG only needs to save the weights of $\mathcal{F}^\prime_\phi$. As we set the intermediate size $p$ of the $\mathcal{F}^\prime_\phi$ to 2, the total number of parameters for the Qwen2.5-1.5B model is $3L(phr + 2p(h+1) + pkr)$ as we configure separate translators for up-proj, down-proj, and gate-proj. This amounts to about 4.04M parameters, storing only 7.71MB (0.04\% of PRAG) in 16-bit precision. 
The reduced storage cost makes it negligible compared to its generalization ability when used in real applications.
\section{Experiments}

\subsection{Experiments Details}
\paragraph{\noindent\textbf{Datasets.}} 
We assess our approach using a variety of benchmark datasets, each specifically crafted to evaluate distinct reasoning abilities, including multi-hop reasoning and commonsense inference. The selected datasets are \textbf{2WikiMultihopQA (2WQA)}~\cite{ho2020constructingmultihopqadataset}, \textbf{HotpotQA (HQA)}~\cite{yang2018hotpotqadatasetdiverseexplainable}, \textbf{PopQA (PQA)}~\cite{mallen2023trustlanguagemodelsinvestigating} and \textbf{ComplexWebQuestions (CWQ)}~\cite{talmor2018webknowledgebaseansweringcomplex}. We provide detailed information about these datasets in Appendix~\ref{app:imple}.

\begin{table*}[t]
\caption{The experimental results of DyPRAG are compared with parametric RAG and standard RAG methods. All metrics are reported as F1 scores (\%). The best performance is bolded, while the second-best is underlined. The \textbf{Avg} is the average performance over all sub-tasks.}
\label{tab:main}
\small
\setlength\tabcolsep{4pt}
\renewcommand{\arraystretch}{1.2}
\centering
\resizebox{\textwidth}{!}{%
\begin{tabular}{@{}lc*{5}{c}*{3}{c}ccc@{}}
\toprule
\multirow{2}{*}{\textbf{Base LLM}} & 
\multirow{2}{*}{\makecell[c]{\textbf{Method}}} & 
\multicolumn{5}{c}{\textbf{2WQA}} & 
\multicolumn{3}{c}{\textbf{HQA}} & 
\multirow{2}{*}{\textbf{PQA}} & 
\multirow{2}{*}{\textbf{CWQ}} &
\multirow{2}{*}{\textbf{Avg}} 

\\
\cmidrule(lr){3-7}\cmidrule(lr){8-10}
 & & 
\textbf{Compare} & \textbf{Bridge} & \textbf{Inference} & \textbf{Compose} & \textbf{Total} & 
\textbf{Bridge} & \textbf{Compare} & \textbf{Total} & & &\\
\midrule
\multirow{6}{*}{\textbf{LLaMA3.2-1B}} & 
\textbf{Vanilla} & 
~42.89 & ~24.17 & ~16.91 & ~7.87 & ~22.52 & 
~13.25 & ~40.26 & ~18.79 & ~2.26 & ~34.94 & ~22.39 \\
&
\textbf{Standard RAG} & 
~41.23 & ~26.78 & ~22.51 & ~\underline{10.21} & ~23.12 & 
~21.38 & ~42.46 & ~\underline{27.14} & ~17.65 & ~37.39 & ~26.99 \\

& \textbf{PRAG} & 
~50.20 & ~24.34 & ~19.11 & ~8.24 & ~27.73 & 
~13.65 & ~40.90 & ~21.50 & ~23.58 & ~35.86 & ~26.51 \\

& \textbf{PRAG-Combine} & 
~40.50 & ~31.30 & ~\textbf{22.85} & ~9.77 & ~\textbf{30.30} & 
~\textbf{22.56} & ~41.55 & ~\textbf{28.31} & ~\textbf{32.59} & ~\textbf{39.63} & ~\underline{29.94} \\

& \textbf{DyPRAG (ours)} & 
~\underline{51.25} & ~\textbf{48.15} & ~17.35 & ~7.54 & ~25.31 & 
~14.05 & ~\textbf{43.90} & ~19.97 & ~11.33 & ~36.86 & ~27.57 \\

& \textbf{DyPRAG-Combine (ours)} & 
~\textbf{52.13} & ~\underline{46.19} & ~\underline{22.54} & ~\textbf{12.60} & ~\underline{29.18} & 
~\underline{22.05} & ~\underline{43.78} & ~26.58 & ~\underline{29.93} & ~\underline{38.96} & ~\textbf{31.80} \\
\midrule
\multirow{6}{*}{\textbf{Qwen2.5-1.5B}} & 
\textbf{Vanilla} & 
~\textbf{45.74} & ~39.06 & ~\underline{17.04} & ~7.27 & ~26.87 & 
~12.18 & ~39.46 & ~17.76 & ~2.87 & ~26.47 &~25.79 \\
&
\textbf{Standard RAG} & 
~38.75 & ~38.84 & ~11.87 & ~5.68 & ~24.31 & 
~16.19 & ~37.13 & ~20.73 & ~9.97 & ~28.23 & ~23.17 \\

& \textbf{PRAG} & 
~\underline{44.96} & ~43.96 & ~\textbf{19.29} & ~\textbf{11.14} & ~\textbf{27.55} & 
~13.27 & ~40.42 & ~18.42 & ~21.55 & ~30.82 & ~\underline{27.14} \\

& \textbf{PRAG-Combine} & 
~40.50 & ~44.00 & ~16.30 & ~8.17 & ~\underline{27.49} & 
~\underline{18.86} & ~36.49 & ~\underline{23.10} & ~\textbf{23.43} & ~\underline{32.13} & ~27.05 \\

& \textbf{DyPRAG (ours)} & ~43.03 & ~\textbf{47.20} & ~\underline{17.04} & ~8.55 & ~26.46
 & 
~13.72 & ~\textbf{41.39} & ~19.67 & ~6.64 & ~31.94 & ~25.56 \\

& \textbf{DyPRAG-Combine (ours)} & 
~35.83 & ~\underline{44.89} & ~14.81 & ~\underline{8.64} & ~25.18
 & 
~\textbf{21.56} & ~\underline{41.25} & ~\textbf{27.57} & ~\underline{22.69} & ~\textbf{33.57} & ~\textbf{27.60} \\
\midrule

\multirow{6}{*}{\textbf{LLaMA3-8B}} & 

\textbf{Vanilla} & 
~54.90 & ~55.20 & ~24.59  & ~14.43 & ~33.02 & 
~19.00& ~45.63 & ~21.29 & ~7.96 & ~\underline{42.44} &~31.85 \\

&
\textbf{Standard RAG} & 
~58.43 & ~47.77 & ~19.20 & ~11.07 & ~34.55 & 
~19.68& ~42.10 & ~24.23 & ~16.13 & ~35.45 & ~30.86 \\

& \textbf{PRAG} & 
~57.78 & ~\underline{58.93} & ~27.61 & ~19.17 & ~39.19 & 
~\underline{33.68} & ~\underline{65.88} & ~38.08 & ~26.13& ~\textbf{43.54}& ~41.00 \\

& \textbf{PRAG-Combine} & 
~\underline{60.13} & ~56.69 & ~\underline{32.71} & ~\underline{20.91} & \underline{40.55} & 
~\textbf{39.41} & ~\textbf{68.22} & ~\textbf{44.84} & ~\underline{26.23} & ~36.41 & ~\underline{42.61} \\

& \textbf{DyPRAG (ours)} & 
~57.39 & ~56.43 & ~25.33 & ~18.88 & ~37.80 & 
~24.85 & ~58.59 & ~28.56 & ~13.60 & ~41.87 & ~36.23 \\

& \textbf{DyPRAG-Combine (ours)} & 
~\textbf{66.00} & ~\textbf{59.46} & ~\textbf{35.78} & ~\textbf{26.90} & \textbf{50.24} & 
33.37 & 57.93 & ~\underline{38.35} & \textbf{32.86} & ~39.07 & ~\textbf{43.69} \\

\bottomrule
\end{tabular}%
}

\end{table*}

\paragraph{\noindent\textbf{Evaluation Metrics.}} 
For evaluation, we use the F1 score (\%), which balances precision and recall by accounting for partially correct answers. Both 2WQA and HQA categorize questions by reasoning type, with 2WQA having four categories and HQA two. To compare DyPRAG with other RAG baselines across reasoning tasks, Table~\ref{tab:main} and~\ref{tab:main_sup} report performance on each sub-task separately, using the first 300 questions from each sub-dataset, following~\cite{su2025parametricretrievalaugmentedgeneration}.

\paragraph{\noindent\textbf{Implementation Details.}}
\label{sec:imple}
To ensure broad effectiveness across models, we select LLMs of varying scales and series, including Qwen2.5-1.5B-Instruct~\citep{qwen2025qwen25technicalreport}, LLaMA-3.2-1B-Instruct~\citep{Llama-3.2-1B-Instruct} and LLaMA-3-8B-Instruct~\citep{Llama-3-8B-Instruct}. For our base experiments, we collect 200 additional questions from each non-overlapping sub-dataset. The number of retrieved documents $c$ is set to 3, resulting in a alignment set $\mathcal{K}$ of 4,800 samples for the parameter translator. The intermediate size $p$ is set to 32. All experiments were conducted using PyTorch on NVIDIA A100 GPUs (80GB) and 3090 GPUs (24GB). Please
refer to Appendix~\ref{app:imple} for more detailed settings.

\subsection{Baselines}

We choose the following RAG baselines for comparison: 

\begin{itemize}
\item  \textbf{Vanilla} represents the answer from original LLMs without any external knowledge.
\item \textbf{Standard RAG}. 
Appends top-retrieved documents to the LLM’s input prompt, explicitly instructing the model to reference them when answering.
\item \textbf{PRAG}  injects relevant documents into the LLM’s parameters via offline parameterization, reducing reliance on input context length.
\item  \textbf{DyPRAG}. Our proposed dynamic parametric RAG which utilize the parameter translator $\mathcal{F}^\prime_{\phi}$ to imitate underlying $\mathcal{F}$. For any retrieved document, we can directly transform it into parametric form without high training and storage cost introduced by PRAG. 
\end{itemize}
Following the approach in~\cite{su2025parametricretrievalaugmentedgeneration}, we conduct experiments that combine in-context injection and parameter injection to explore their interaction. This results in two additional baselines, referred to as \textbf{PRAG-Combine} and \textbf{DyPRAG-Combine}.

\subsection{Main Results}

In this section, we present the main experimental result and in-depth analysis of DyPRAG compared with vanilla model, standard RAG,  efficient RAG, PRAG baselines and a combined setting
% that leverages both parametric and contextual knowledge
. Notably, the vanilla model occasionally outperforms RAG in certain situations. We analyze the reasons for this in Appendix~\ref{app:rag_deficiency}, and confirm that it won't affect the subsequent analysis.
\paragraph{\noindent\textbf{Overall Analysis.}} Since PRAG learns the mapping function $\mathcal{F}$ by training separately for each document, it can be considered as a performance upper bound of DyPRAG. 
Remarkably, our proposed DyPRAG achieves comparable or even superior results across various tasks as shown in Table~\ref{tab:main}. For instance, using LLaMA3.2-1B, DyPRAG achieves an average score of 27.57\%, surpassing PRAG by 1.06\%, standard RAG by 0.58\% and vanilla model by 5.18\%.
A notable result is observed in the bridge sub-task of 2WQA, where DyPRAG achieves a score of 48.15\%, outperforming RAG and PRAG by a significant margin of 21.37\% and 23.81\%, respectively. 
This demonstrates that our method learns more useful information when trained on diverse datasets. 
We also compare our DyPRAG framework with several efficient RAG baselines, including FLARE~\cite{jiang2023active} and DRAGIN~\cite{su2024dragin}. As shown in Table~\ref{tab:main_sup}, both DRAGIN and FLARE outperform standard RAG in most 2WQA settings. However, DyPRAG achieves even better results, demonstrating its superiority. For example, when using LLaMA3-8B as the base model, DyPRAG outperforms DRAGIN and FLARE by 1.56\% and 2.63\% on 2WQA, respectively. These results highlight the consistent performance improvements offered by DyPRAG over other efficient RAG methods, underscoring its effectiveness for test-time knowledge enhancement.

% \paragraph{\noindent\textbf{Performance deficiencies.}} In experiments with Qwen2.5-1.5B (LLaMA3.2-8B), our DyPRAG averagely outperforms standard RAG by 1.10\% (5.77\%), but lags behind the upper-bound PRAG by 2.87\% (4.77\%). The performance gap primarily lies in the PQA task, where DyPRAG struggles to optimize due to the factual nature of the questions, which are predominantly occupation-related. The results indicate that single DyPRAG encounters challenges when processing excessively fine-grained knowledge. However, it still outperforms standard RAG significantly, demonstrating the efficiency of parameter injection at test-time.
\begin{table}
    \begin{minipage}[t]{0.48\textwidth}
        \caption{The OOD performance on two open-domain datasets for $\mathcal{F}^\prime_\phi$ trained on $\mathcal{K}$ is reported.}
\label{tab:ood_performance}
\small
\centering
\resizebox{0.83\linewidth}{!}{
\begin{tabular}{@{}lccc@{}}
\toprule
\textbf{Base Model} & \textbf{Method} & \textbf{IIRC} & \textbf{SQA} \\ 
\midrule
\multirow{4}{*}{\textbf{LLaMA3.2-1B}} 
 & \textbf{Vanilla} & 10.99 & 21.67\\ 
 & \textbf{Standard RAG} & 40.38 & 27.67 \\ 
 & \textbf{DyPRAG} & 14.04 & 39.67\\
 & \textbf{DyPRAG-Combine} & \textbf{41.91} & \textbf{50.33} \\ 
\midrule
\multirow{4}{*}{\textbf{Qwen2.5-1.5B}}
 & \textbf{Vanilla} & 8.78 & 1.00 \\ 
 & \textbf{Standard RAG} & 30.52 & 39.00 \\ 
 & \textbf{DyPRAG} & 10.23 & 15.67\\
 & \textbf{DyPRAG-Combine} & \textbf{38.25} & \textbf{43.33} \\ 
 \midrule
 \multirow{4}{*}{\textbf{LLaMA3-8B}} 
 & \textbf{Vanilla} & 13.23 & 33.33 \\  
 & \textbf{Standard RAG} &  43.27 & 45.67 \\
  & \textbf{DyPRAG} & 18.16 & 45.67\\
 & \textbf{DyPRAG-Combine} & \textbf{57.90} & \textbf{58.67} \\ 
 
\bottomrule
\end{tabular}
}
    \end{minipage}
    \hfill
    \begin{minipage}[t]{0.48\textwidth}
      \caption{Ablation study of intermediate dimension $p$ of $\mathcal{F}^\prime_\phi$. The backbone model is the Qwen2.5-1.5B. The inference time is computed by average time of CWQ with batch\_size of 1. The encode time is highlighted in \textcolor{red}{red}, while the translate time is marked in \textcolor{blue}{blue}.
        }
    \label{tab:inter_dim_p}
        \centering
        \small
\resizebox{1.0\textwidth}{!}{%
        \begin{tabular}{p{2.5cm}ccc}
            \toprule
            Method & \begin{tabular}[c]{@{}c@{}}CWQ\\F1\end{tabular} & \begin{tabular}[c]{@{}c@{}}Inference Time\\ (s) \end{tabular} & \begin{tabular}[c]{@{}c@{}}Storage Cost \\ (MB)\end{tabular} \\
            \midrule
            Vanilla & \cellcolor{blue!5}26.47 & \cellcolor{green!20}0.56 (0.47x) & - \\ 
            RAG & \cellcolor{blue!10}28.32 & \cellcolor{green!50}1.20 (1x) & - \\
            PRAG & \cellcolor{blue!20}30.82 & \cellcolor{green!20}0.56 (0.47x) & \cellcolor{red!40}672 (1x) \\
            \midrule
            DyPRAG ($p=2$) & \cellcolor{blue!30}32.66 & \cellcolor{green!25}0.56+\textcolor{red}{0.13}+\textcolor{blue}{0.060} (0.625x) & \cellcolor{red!5}7.71 (0.011x) \\
            DyPRAG ($p=4$) & \cellcolor{blue!40}33.26 & \cellcolor{green!27}0.56+\textcolor{red}{0.13}+\textcolor{blue}{0.062} (0.627x) & \cellcolor{red!10}15.42 (0.023x) \\
            DyPRAG ($p=16$) & \cellcolor{blue!25}32.08 & \cellcolor{green!23}0.56+\textcolor{red}{0.13}+\textcolor{blue}{0.055} (0.621x) & \cellcolor{red!20}61.70 (0.092x) \\
            DyPRAG ($p=32$) & \cellcolor{blue!20}31.94 & \cellcolor{green!25}0.56+\textcolor{red}{0.13}+\textcolor{blue}{0.060} (0.625x) & \cellcolor{red!30}123.39 (0.184x) \\
            \bottomrule
        \end{tabular}
        }
    \end{minipage}
\end{table}

\paragraph{\noindent\textbf{DyPRAG-Combine Leads to Superior Performance.}} By combining in-context injection with parameter injection, DyPRAG-Combine achieves the best performance across all models, outperforming all baselines. For instance, DyPRAG-Combine outperforms PRAG-Combine by 1.86\% on LLaMA3.2-1B, 0.46\% on Qwen2.5-1.5B and 1.08\% on LLaMA3-8B on average. These results demonstrate that the dynamic parameters introduced by our approach effectively intergrade with contextual knowledge, enabling the two information sources to complement each other.

\subsection{Out-Of-Distribution (OOD) Performance}
To further demonstrate the generalization ability of the DyPRAG method, we evaluate it in an out-of-distribution (OOD) scenario. Table~\ref{tab:ood_performance} presents the OOD performance on the StrategyQA (\textbf{SQA})\cite{geva2021didaristotleuselaptop} and \textbf{IIRC}\cite{ferguson-etal-2020-iirc} datasets. Compared to the standard RAG model, the vanilla model performs poorly due to the lack of relevant knowledge. In contrast, our DyPRAG successfully injects parametric knowledge, achieving comparable performance in SQA. Notably, DyPRAG-Combine which incorporates document-related parametric knowledge with ground-truth passages can answer these questions more correctly, achieving the best performance across all scenarios. For example, DyPRAG-Combine improves performance on SQA by 22.66\% using LLaMA3.2-1B and on IIRC by 13.63\% using LLaMA3-8B. These experimental results highlight the significant generalization ability of DyPRAG, surpassing that of PRAG, which is unable to handle this OOD scenario without additional offline training. Detailed implementation is provided in Appendix~\ref{para:ood_imple} and analysis in Appendix~\ref{app:rag_deficiency}.

\subsection{Ablation Study}
\label{sec:ablation}
\paragraph{\noindent\textbf{Effect of Intermediate Dimension $p$.}}As demonstrated in Section~\ref{sec:cost_ana}, the total storage cost for $\mathcal{F}^\prime_\phi$ is $3L(phr+2p(h+1)+pkr)$, which scales linearly with $p$. Therefore, we conducted an ablation study on $p$. 
As shown in Table~\ref{tab:inter_dim_p}, DyPRAG consistently outperforms both standard RAG and PRAG. Surprisingly, $p=2$ achieves the second-best performance with a storage cost of only 7.71MB. Notably, our DyPRAG significantly reduces inference costs compared to standard RAG, while introducing only minimal overhead (i.e., encode and translate processes) compared to PRAG.
In contrast, PRAG requires 672MB to store data for 300 test questions in CWQ, resulting in a significant overhead.
The experiments demonstrate that our proposed DyPRAG not only drastically reduces storage costs but also enhances task performance, showcasing exceptional robustness.

\subsection{Analysis of Contextual and Parametric Knowledge Conflict and Fusion}
When contextual knowledge and parametric knowledge conflict, LLMs may struggle to determine which source of information is more reliable~\cite{tao2024contextleadsparametricmemory,zhang2024evaluatingexternalparametricknowledge}. A key distinction between parametric RAG and standard RAG lies in whether external documents are used as contextual knowledge or transforming into parametric knowledge. Additionally, we aim to explore why combining parameter injection with in-context injection results in superior performance.

\begin{wraptable}{r}{0.5\textwidth}
\caption{Case study about contextual and parametric knowledge conflict in 2WQA where only DyPRAG-Combine answers correctly (11.33\%). The backbone model is the LLaMA3.2-1B.  \colorbox{red!30}{\textcolor{red!30}{1}}: deficiency in parametric knowledge,
\colorbox{yellow!60}{\textcolor{yellow!60}{1}}: knowledge conflict,
\colorbox{green!30}{\textcolor{green!30}{1}}: successful knowledge manipulation.}
\label{tab:conflict_case_1}
\centering
\small
\resizebox{0.5\textwidth}{!}{%
\begin{tabular}{lcc}
\hline
\multicolumn{3}{p{0.95\linewidth}}{\textbf{Question:} Which film whose director was born first, \colorbox{yellow!60}{The Snake Brothers} or \colorbox{green!30}{Olympus Has Fallen}?} \\
\hline
\multicolumn{3}{p{0.95\linewidth}}{\textbf{Ground truth:} \colorbox{green!30}{Behind Prison Gates}} \\
\hline
\multicolumn{3}{p{0.95\linewidth}}{\textbf{Retrieved top-1 document:} Roman Waugh was announced as director for the film. \colorbox{green!30}{Olympus Has Fallen} (film series) The Olympus Has Fallen film series includes action thriller movies...} \\
\hline
\textbf{Method} & \textbf{Answer} & \textbf{Status} \\
\hline
\textbf{Vanilla} & \colorbox{red!30}{David R} & \textcolor{red}{\textsf{\XSolidBrush}} \\
\hline
\textbf{Standard RAG} & \colorbox{yellow!60}{The Snake Brothers} & \textcolor{red}{\textsf{\XSolidBrush}} \\
\hline
\textbf{DyPRAG} (ours) & \colorbox{red!30}{The Snake Brothers} & \textcolor{red}{\textsf{\XSolidBrush}} \\
\hline
\textbf{DyPRAG-Combine} (ours)  & \colorbox{green!30}{Olympus Has Fallen} & \textcolor{green}{\Checkmark}  \\
\hline
\end{tabular}
}

\end{wraptable}

% According to experimental results, DyPRAG achieves significantly bettser performance in the Bridge sub-task of 2WQA using LLaMA-1B as the base model. 
\paragraph{\noindent\textbf{Contextual + Parametric Knowledge Relieves RAG Hallucination.}} As shown in Table~\ref{tab:conflict_case_1}, the vanilla LLM contains incorrect parametric knowledge.  While standard RAG and DyPRAG try to address this by performing in-context injection and parameter injection, respectively. Both approaches introduce hallucinations that result in incorrect answers. 

In contrast, DyPRAG-Combine effectively integrates contextual knowledge with transformed parametric knowledge, enabling it to provide correct answers and demonstrating its ability to leverage both types of knowledge effectively.
Compared to  RAG, DyPRAG-Combine transforms the retrieved documents into parametric knowledge before injecting them into the model. This approach ensures that the LLM already contains relevant knowledge when answering questions, unlike RAG which often suffers from well-known hallucination issues~\cite{tao2024contextleadsparametricmemory,sun2025redeepdetectinghallucinationretrievalaugmented}. As a result, DyPRAG-Combine offers a practical and efficient solution to directly relieve the common hallucination problem in standard RAG, providing a plug-and-play improvement to enhance RAG's performance in real-world applications. 

\begin{wrapfigure}{r}{0.5\textwidth}
\centerline{\includegraphics[width=0.5\textwidth]{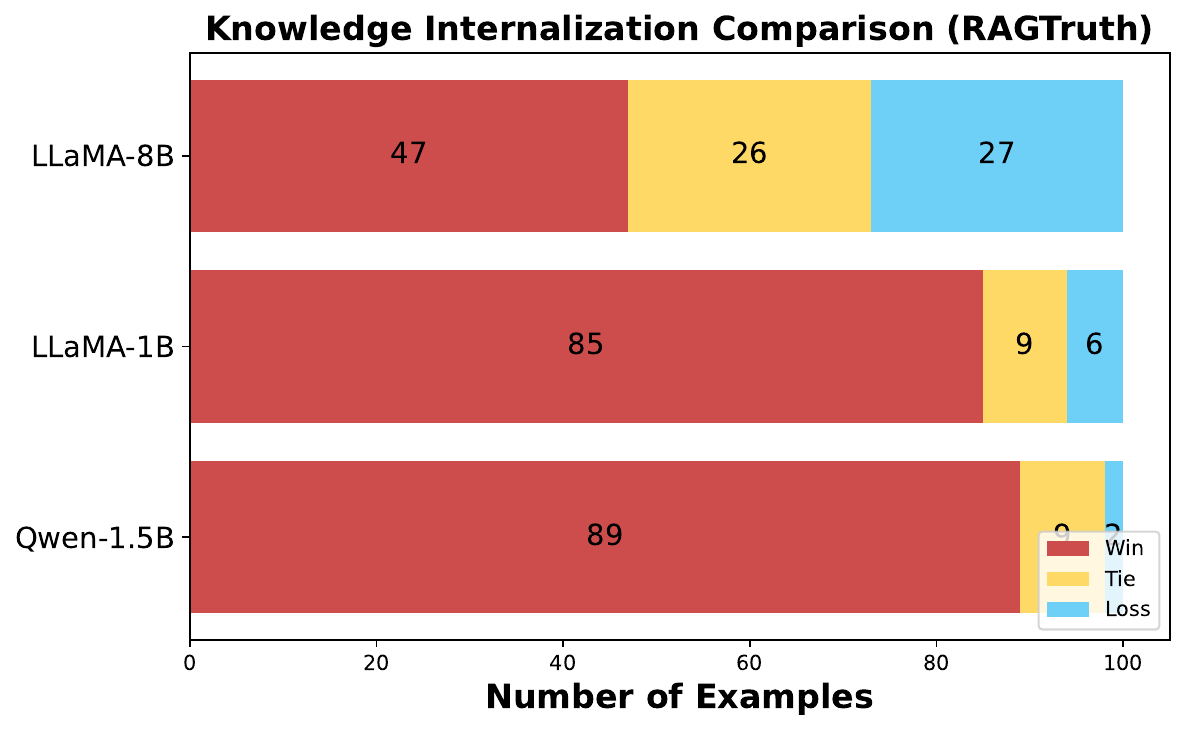}}
\caption{
Comparison between DyPRAG-Combine vs standard RAG judged by GPT-4o. 
}
\label{fig:ragtruth}
% \vspace{0.5cm}
\end{wrapfigure}

\paragraph{\noindent\textbf{DyPRAG Enables LLMs to Internalize Unseen Knowledge.}} 
The retrieved documents in our experiments are primarily sourced from Wikipedia, which are already encountered by LLMs during the pre-training phase. In this section, we further investigate how DyPRAG performs on unseen contextual knowledge using the RAGTruth benchmark~\cite{niu2023ragtruth}. Specifically, we randomly sample 100 examples from the QA-type sub-dataset, which presents greater challenges (e.g., the required answers are only accessible in carefully crafted context). 
As shown in Figure~\ref{fig:ragtruth}, DyPRAG-Combine significantly outperforms RAG. This demonstrates that DyPRAG effectively enables LLMs to better internalize contextual knowledge and mitigate hallucinations, even when handling unseen data.  \footnote{We explore metrics for detecting hallucinations in RAG in Appendix~\ref{app:rag_hal}. Additionally, we present a detailed analysis of contextual and parametric knowledge in Appendix~\ref{app:context_param_ana}.}

\section{Conclusion}
In this work, we propose Dynamic Parametric RAG (DyPRAG), a novel framework designed to reduce the high inference cost of standard RAG while also mitigating the substantial training and storage costs of previous Parametric RAG approach. DyPRAG learns the underlying mapping function from documents to parameters, enabling effective parameter injection in a plug-and-play manner at test time, thereby enhancing the internal knowledge of LLMs .
Through comprehensive experiments, we demonstrate that DyPRAG achieves superior performance with significantly lower costs and great generalization ability. Notably, this enables the discovery of a powerful RAG paradigm, where DyPRAG effectively combines parametric knowledge with contextual knowledge. The superior knowledge fusion ability comes with extremely low costs while mitigating RAG hallucinations,  revealing its practical capacity in real-world RAG systems.

%%%%%%%%%%%%%%%%%%%%%%%%%%%%%%%%%%%%%%%%%%%%%%%%%%%%%%%%%%%%
\bibliography{custom}
\bibliographystyle{plain}
\appendix
\newpage
\begin{table}[t]
\centering
\caption{Comparison of cost metrics for different baselines. \text{ATTN} denotes the time complexity of the self-attention module as $O(|I|^2h)$, and \text{FFN} represents the FFN with $O(|I|h^2)$, where context length $|I|=1$ and $|R|$ denotes the response length. \colorbox{red!60}{\textcolor{red!60}{ii}} indicates significant high cost, \underline{\phantom{aa}} denotes negligible cost, and \cancel{\phantom{aa}} represents temporal storage.}
\label{tab:comparison}
\resizebox{\textwidth}{!}{
\begin{tabular}{lccc}
\toprule
\textbf{Method} & \textbf{Inference Cost} & \textbf{Training Cost} & \textbf{Storage Cost} \\
\midrule
\textbf{RAG} & \colorbox{red!60}{$|R|\times((c|d|+|q|)^2\times\text{ATTN}+(c|d|+|q|)\times\text{FFN})$} & - & - \\
\hline
\textbf{PRAG} & $|R|\times(|q|^2\times\text{ATTN}+|q|\times\text{FFN}) $ 
&

$\begin{array}{c}
M\times(9|d|^2\times\text{ATTN}+3|d|\times\text{FFN})+\\
\colorbox{red!60}{$M\times E_1\times(81|d|^2\times\text{ATTN}+9|d|\times\text{FFN})$}
\end{array}
$
& \colorbox{red!60}{$M\times6Lr(h+k)$} \\
\hline
\textbf{DyPRAG} & $\begin{array}{c}
c\times(|d|^2\times\text{ATTN}+|d|\times\text{FFN})+\\\underline{c\times O(p(h+1+hr)})+\\
|R|\times(|q|^2\times\text{ATTN}+|q|\times\text{FFN})
\end{array}$ 
& $\begin{array}{c}
N\times(9|d|^2\times\text{ATTN}+3|d|\times\text{FFN})+\\
N\times E_1\times(81|d|^2\times\text{ATTN}+9|d|*\text{FFN})+\\N\times E_2\times(9(|qa|+|d|)^2\times\text{ATTN}+3(|qa|+|d|)\times\text{FFN})+\underline{ O(p(h+1+hr)})\end{array}$ & 
$\begin{array}{c}
     \cancel{N*6Lr(h+k)}+\\
     3L(phr+2p(h+1)+pkr) 
\end{array}$ \\
\bottomrule
\end{tabular}%
}
\end{table}

\section{Detailed Cost Comparison}
\label{app:detail_compare}
In this section, we provide a detail comparison of several cost metrics for standard RAG, PRAG and our proposed DyPRAG, as shown in Table~\ref{tab:comparison}.
\paragraph{\noindent\textbf{Inference Cost.}}We first analyze the inference cost across three baselines. Intuitively, the RAG method requires more resources for inference due to its context length of $c|d| + |q|$, compared to only $|q|$ for PRAG and DyPRAG. In our experimental settings, $|q|$ is usually less than 100, while $|d|$ is typically larger than 600, with $c$ set to 3. This results in an attention cost of at least 271x and a FFN cost of 19x for RAG. 
For DyPRAG, there is additional cost incurred for encoding and translating. The encoding cost is $c \times (|d|^2 \times \text{ATTN} + |d| \times \text{FFN})$, as each document should be encoded separately. As shown in Table~\ref{tab:inter_dim_p}, the encoding time is significantly lower than the inference time because encoding requires only a single forward pass. Additionally, the translation time is also negligible. Moreover, the response length $|R|$ exhibits a linear relationship with the LLM inference loss. As illustrated in Figure~\ref{fig:qwen_res_length}, the response length decreases when DyPRAG is employed, enabling LLMs to better internalize knowledge. Notably, DyPRAG-Combine achieves much shorter response lengths, significantly reducing inference costs compared to standard RAG.

\paragraph{\noindent\textbf{Training Cost.}} PRAG~\cite{su2025parametricretrievalaugmentedgeneration} introduces further training for each document to obtain corresponding LoRA parameters. In Section~\ref{sec:cost_ana}, we hypothesize that after augmentation, there are a total of $3|d|$ tokens, resulting in a cost of $N \times (9|d|^2 \times \text{ATTN} + 3|d| \times \text{FFN})$ for DyPRAG and $M \times (9|d|^2 \times \text{ATTN} + 3|d| \times \text{FFN})$ for PRAG, where $N$ represents the size of the training dataset $\mathcal{K}$ and $M$ denotes the size of the test set. The common divisor of offline parametrization is $E_1 \times (81|d|^2 \times \text{ATTN} + 9|d| \times \text{FFN})$, where $E_1$ is the number of epochs for LoRA training.

Additionally, to train our $\mathcal{F}^\prime_\phi$ for $E_2$ epochs, we need to perform both forward and backward passes (the backward pass requires twice the cost of the forward pass) on one QA pair and its corresponding document in each step. This results in a cost of $N \times E_2 \times 9(|qa| + |d|)^2 \times \text{ATTN} + 3(|qa| + |d|) \times \text{FFN}$, with a negligible cost for translation. As shown in Figure~\ref{fig:llama_vary} and ~\ref{fig:qwen_vary}, our DyPRAG achieves stable results with as few as 480 examples (even fewer is powerful), while $M = 3000$ in our experiments, and this value would be significantly larger in real-world applications.

For instance, using LLaMA3-8B as the backbone, producing a $p_i$ requires 88 seconds, while one step for $\mathcal{F}^\prime_{\phi}$ only takes an average of 15 seconds. Therefore, the total cost for training (excluding augmentation) is $M \times 88s$ in PRAG and $N \times 103s$ in DyPRAG. Assuming $N = 480$ and $M = 3000$, DyPRAG is 5.34x faster than PRAG. The low requirement for a large $N$ makes DyPRAG highly effective and generalizable for real-world scenarios, with extremely low costs that can be handled during offline training.

\paragraph{\noindent\textbf{Storage Cost.}} As illustrated in Section~\ref{sec:cost_ana}, each $p_i$ requires 6.73MB for PRAG using Qwen2.5-1.5B, resulting in a total storage cost of 18.66GB in our main experiment. However, we significantly reduce this cost by imitating the underlying function between the document and parameters. Notably, the cost for $p_i$ is a temporary cost in DyPRAG, which can be removed after collecting data or training one $p_i$ and then updating $\mathcal{F}^\prime_\phi$ by one step. Consequently, the overall cost of DyPRAG is substantially lower than that of PRAG (e.g., DyPRAG achieve better performance with only 7.71MB of storage as shown in Table~\ref{tab:inter_dim_p}).

\section{Experiment Setup}
\subsection{Implementation Details}
\label{app:imple}
\paragraph{\noindent\textbf{QA Datasets.}} 
To ensure a comprehensive evaluation, we assess our method using the following datasets:
\begin{itemize}

\item\textbf{2WikiMultihopQA (2WQA)}~\cite{ho2020constructingmultihopqadataset} is designed to evaluate a model’s capability in multi-hop reasoning by synthesizing information from multiple Wikipedia passages.

\item\textbf{HotpotQA (HQA)}~\cite{yang2018hotpotqadatasetdiverseexplainable} similarly targets multi-hop reasoning, requiring models to amalgamate information from various contexts to answer a single query.

\item\textbf{PopQA (PQA)}~\cite{mallen2023trustlanguagemodelsinvestigating} focuses on factual question answering, posing challenges that test the model’s ability to recall precise knowledge and navigate ambiguities in entity representation.

\item\textbf{ComplexWebQuestions (CWQ)}~\cite{talmor2018webknowledgebaseansweringcomplex} entails answering complex, multi-step questions sourced from the web, further challenging the model’s capacity to retrieve and reason over extensive web content.
\end{itemize}

\paragraph{\noindent\textbf{Offline Doc-Param Pairs Collection.}}
Following~\cite{jiang2023active, su2025parametricretrievalaugmentedgeneration}, we utilize Wikipedia dumps as the external knowledge corpus, adopting the dataset proposed by DPR~\cite{karpukhin2020dense}.
For document augmentation, each document is rewritten once, and three QA pairs are generated based on the document. Unless explicitly stated otherwise, the downstream LLM is used for this purpose.
During LoRA fine-tuning, the learning rate was set to $3 \times 10^{-4}$, and training was conducted for a single epoch (except PQA for 2). The LoRA modules were integrated exclusively into the feed-forward network (FFN) matrices, while the query, key, and value (QKV) matrices were excluded. The scaling factor $\alpha$ was set to 32, the LoRA rank $r$ was configured to 2, and no dropout was applied to ensure training stability and maximize parameter updates.
The LoRA weights were randomly initialized following the settings outlined in the original LoRA paper~\cite{hu2021loralowrankadaptationlarge}.

\paragraph{\noindent\textbf{Inference Settings.}}
All experiments use the publicly available Hugging Face implementations of LLaMA and Qwen. To ensure fairness, DyPRAG and all baselines share the same prompt template in Figure  ~\ref{fig:nocot_format} and ~\ref{fig:cot_format} following~\cite{su2025parametricretrievalaugmentedgeneration} and adopt of greedy decoding for result reproducibility. The max number of new tokens is set to 128.

\paragraph{\noindent\textbf{Retrieval Module $\mathcal{R}$.}}
Recent research on retrieval-augmented generation (RAG)~\cite{ram2023incontextretrievalaugmentedlanguagemodels} has shown that BM25 matches or even surpasses state-of-the-art dense retrieval models in certain scenarios. Following~\cite{su2025parametricretrievalaugmentedgeneration}, we adopt BM25 as the retriever in our approach and Elasticsearch is used as the backend for implementing BM25.

\paragraph{\noindent\textbf{Training $\mathcal{F}^\prime_\phi$.}}
Motivated by~\cite{liao2025instancetraininginstructionlearning}, we use simple MLP hypernetwork to transform embedding into adapter parameters.  Through cross validation, the learning rate was set to $1 \times10^{-5}$, and the training epoch was set to 1 which making the overall alignment process quickly. The truncation max length of text is set to 3000, which is larger than most retrieved documents. The performance reports for Qwen2.5-1.5B and LLaMA3.2-1B in Table~\ref{tab:main} are based on training with 4,800 examples, while LLaMA3-8B is trained on 2,400 examples (except for 480 examples on 2WQA).

\paragraph{\noindent\textbf{Implementation of OOD Experiment.}}
\label{para:ood_imple}
To evaluate the generalization ability of our proposed DyPRAG, we select to out-of-distribution (OOD) datasets to conduct.
\begin{itemize}
    \item \textbf{StrategyQA (SQA)}~\cite{geva2021didaristotleuselaptop}:  A QA benchmark where reasoning steps are implicit in the question and must be inferred through strategic reasoning.
    \item  \textbf{IIRC}~\cite{ferguson-etal-2020-iirc}: A dataset comprising over 13,000 questions based on English Wikipedia paragraphs that provide only partial information, requiring retrieval of missing details from linked documents.
\end{itemize}

For each dataset, we select the first 300 examples for testing and evaluate performance using F1 score for IIRC and Accuracy for SQA as metrics. Both datasets provide with ground-truth passages which indicate a more rigorous evaluation setting. For IIRC, we adopt the few-shot prompts from~\cite{su2024dragin}, while SQA is evaluated in a zero-shot setting. Notably, the same prompt format (in Figure~\ref{fig:nocot_format} and ~\ref{fig:cot_format}) from the main experiment is used to ensure a fair comparison.

\paragraph{\noindent\textbf{Implementation of RAGTruth Experiment.}}RAGTruth~\cite{niu2023ragtruth} is a benchmark dataset designed to evaluate the extent of hallucination in models. For our evaluation, we randomly select 100 QA-type subsets from RAGTruth, ensuring alignment with the training data of $\mathcal{F}^\prime_\phi$. Notably, some questions in RAGTruth require the provided documents to be answerable which are more difficult. Interestingly, during evaluation, we observe that $\mathcal{F}^\prime_\phi$ with fewer trained parameters perform better in such scenarios. Specifically, we train only 480 examples for LLaMA3.2-1B and Qwen2.5-1.5B, and 240 examples for LLaMA3-8B. We use GPT-40 as judge using prompt template in Figure~\ref{fig:ragtruth_format}.

\begin{table}[t]
\caption{The experimental results of DyPRAG are compared with other effective RAG methods. All metrics are reported as F1 scores (\%). The best performance is bolded, while the second-best is underlined. The evaluation is conducted on 2WQA and HQA datasets, focusing exclusively on the total sub-task.}
\label{tab:main_sup}
\centering
\small
\resizebox{0.6\textwidth}{!}{%
\begin{tabular}{@{}lc*{4}{c}@{}}
\toprule
\multirow{2}{*}{\textbf{Base LLM}} & 
\multirow{2}{*}{\makecell[c]{\textbf{Method}}} & 
\multicolumn{1}{c}{\textbf{2WQA}} & 
\multicolumn{1}{c}{\textbf{HQA}} & \multirow{2}{*}{\textbf{Avg}}

\\
\cmidrule(lr){3-3}
\cmidrule(lr){4-4}

& & \textbf{Total} & 
 \textbf{Total} \\
\midrule
\multirow{5}{*}{\textbf{LLaMA3.2-1B}} & 
\textbf{Standard RAG} & 
 ~23.12 & ~\textbf{27.14} & \underline{25.13} \\
& \textbf{DRAGIN} & 21.73 & 12.50 & 17.12\\
& \textbf{FLARE} & 21.55 & 19.38 & 20.47\\
& \textbf{DyPRAG (ours)} & 
 ~\underline{25.31} & 
 ~19.97 & 22.64 \\
& \textbf{DyPRAG-Combine (ours)} & 
~\textbf{29.18} & 
 ~\underline{26.58}  & \textbf{27.88}\\
\midrule

\multirow{5}{*}{\textbf{Qwen2.5-1.5B}} & 
\textbf{Standard RAG} & 
 ~24.31 &  \underline{20.73} &
 ~\underline{22.52}  \\
& \textbf{DRAGIN} & 25.01& 8.51 &16.76\\
& \textbf{FLARE} & 21.56 & 7.97 & 14.77 \\
& \textbf{DyPRAG (ours)} & 
 ~\textbf{26.46} & 
 ~19.67 &23.07 \\
& \textbf{DyPRAG-Combine (ours)} & 
  ~\underline{25.18}& 
 ~\textbf{27.57}  & \textbf{26.38}\\
\midrule
\multirow{5}{*}{\textbf{LLaMA3-8B}} & 

\textbf{Standard RAG} &  ~34.55 & 
 ~24.23 &29.39  \\

& \textbf{DRAGIN} & 35.69& 12.16 & 23.93\\
& \textbf{FLARE} & 34.62 & \underline{29.43} & \underline{32.03}\\

& \textbf{DyPRAG (ours)} & 
 ~\underline{37.25} & ~22.55 & 29.90\\

& \textbf{DyPRAG-Combine (ours)} & 
 \textbf{45.17} & 
 ~\textbf{38.35} &  \textbf{41.76}\\
\bottomrule
\end{tabular}%
}
\end{table}

\section{Supplement Experiment Results}
\label{app:sup}

\paragraph{\noindent\textbf{Comparison with effective RAG baselines.}}
To compare our DyPRAG with effective RAG methods, we introduce two powerful baselines:
\begin{itemize}
    \item FLARE~\cite{jiang2023active} is a multi-round retrieval augmentation method that triggers retrieval whenever it encounters an uncertain token. The query is defined as the last generated sentence excluding the uncertain tokens.
    \item DRAGIN~~\cite{su2024dragin} improves multi-round retrieval by triggering only when an uncertain token has semantic significance and strongly influences subsequent tokens. It formulates queries using the model’s internal state and preceding context.
\end{itemize}

The experimental results are presented in Table~\ref{tab:main_sup}. Compared to standard RAG, DRAGIN and FLARE do not demonstrate significant performance advantages when the model size is smaller (e.g., LLaMA3.2-1B and Qwen2.5-1.5B). However, as the model size increases (e.g., LLaMA3-8B), DRAGIN achieves the best performance on the 2WQA dataset, while FLARE performs best on the HQA dataset comparing with RAG baseline. This indicates that effective RAG methods are often constrained by the model's inherent capabilities and lack robust generalization. In contrast, our proposed DyPRAG consistently delivers superior performance in most cases, demonstrating the effectiveness of our approach. Furthermore, when combined with in-context injection, DyPRAG achieves an average improvement of 6.54\% over standard RAG, highlighting the tremendous potential of integrating parametric knowledge with contextual knowledge.

\begin{figure}[t]
\centerline{\includegraphics[width=\textwidth]{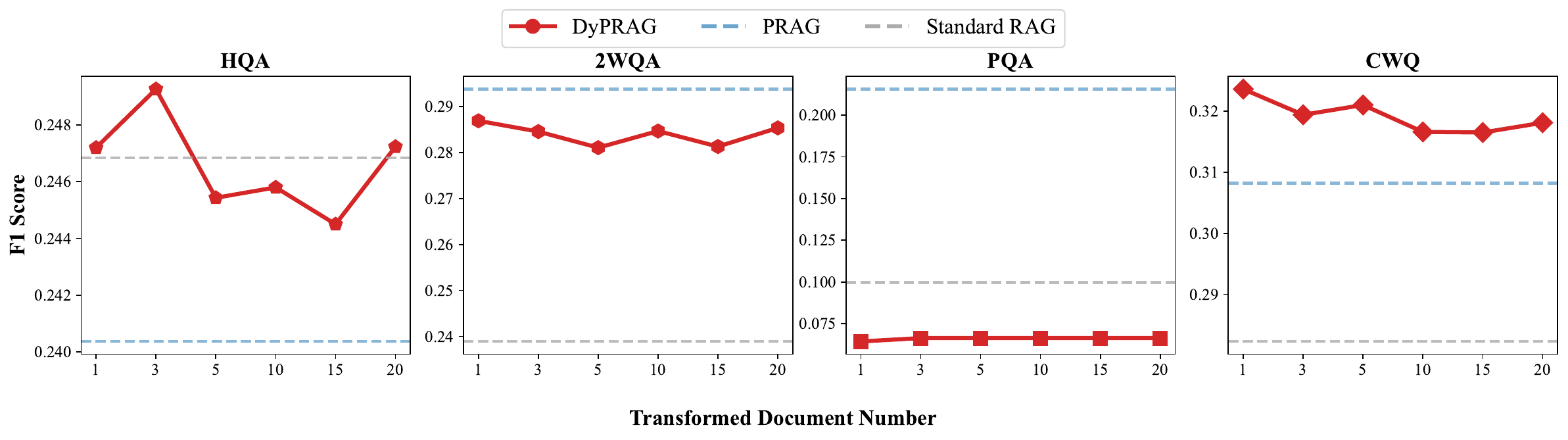}}
\caption{
Performance of Qwen2.5-1.5B with varying number of injected documents.
}
\label{fig:qwen_vary_doc}

\end{figure}
\paragraph{\textbf{\noindent Varying Number of Injected Documents}}
For standard RAG, the number of retrieved documents, denoted as $c$, is a crucial hyperparameter to tune. Recent studies~\cite{leng2024longcontextragperformance, wei2025instructraginstructingretrievalaugmentedgeneration} have investigated the impact of longer context lengths on standard RAG. However, the effect of the number of injected documents in parametric form remains underexplored. Our proposed DyPRAG framework can seamlessly adapt to this scenario due to its inherent flexibility.

As shown in Figure~\ref{fig:qwen_vary_doc}, the performance of DyPRAG does not significantly improve as the number of injected documents increases. For instance, in the 2WQA and CWQ datasets, the best performance is achieved when using only the top-1 document. This indicates that the most relevant document, as determined by the retriever $\mathcal{R}$, is sufficient to provide the knowledge needed to answer the question effectively.
On the other hand, in datasets such as HQA and PQA, the best performance is observed when $c = 3$, suggesting that when more relevant information is retrieved, simple averaging of LoRA parameters can effectively integrate the knowledge. Additionally, in three out of four datasets (except PQA), the model's performance declines when too many documents are injected. This observation aligns with the findings in~\cite{shi2023largelanguagemodelseasily}, which suggest that task-irrelevant redundant information can degrade the model's performance, especially the compression of documents is lossy.

\begin{figure}[t]
    \centering
\begin{minipage}{0.45\textwidth}
    \centering
    \includegraphics[width=\textwidth]{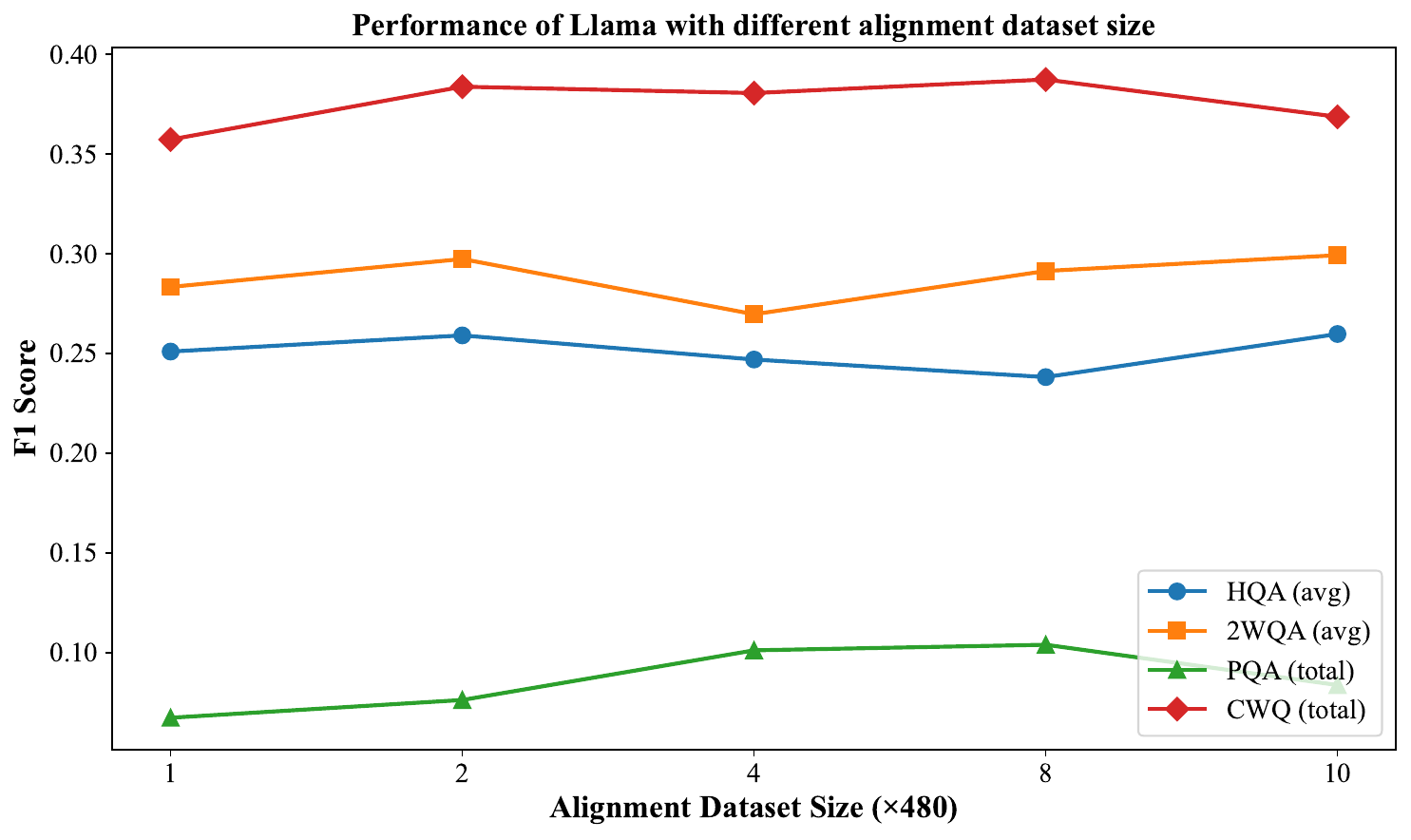}
    % \vspace{-0.45cm}
    \caption{
    Ablation study of varying training dataset size for DyPRAG. The backbone model is the LLaMA3.2-1B.
    }
\label{fig:llama_vary}
\end{minipage}
\hfill
\begin{minipage}{0.48\textwidth}
    \centering
    \includegraphics[width=\textwidth]{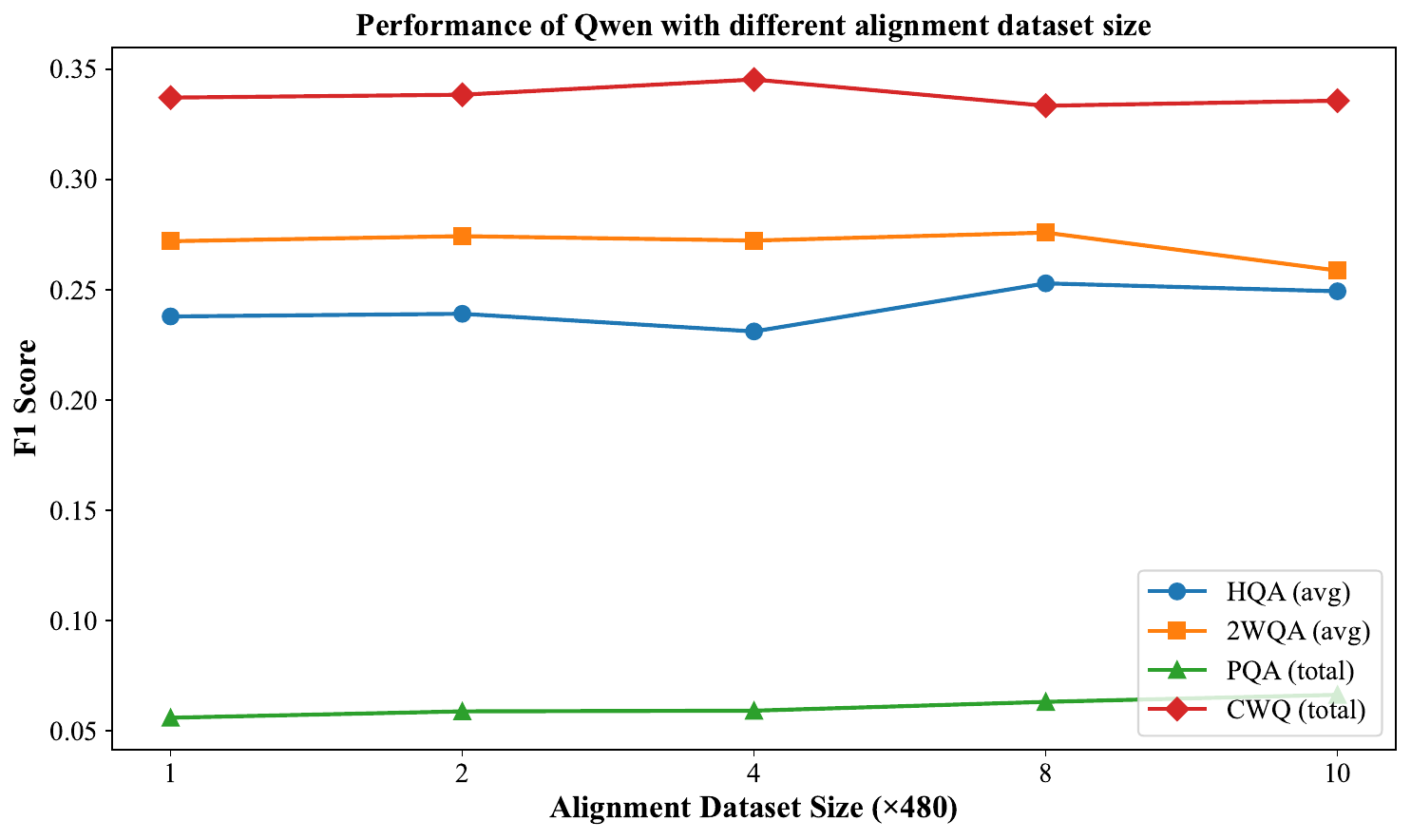}
% \vspace{-0.45cm}
    \caption{
    Ablation study of varying training dataset size for DyPRAG. The backbone model is the Qwen2.5-1.5B.
    }
\label{fig:qwen_vary}
\end{minipage}
\end{figure}
\paragraph{\noindent\textbf{Effect of Doc-Param Pairs Size.}}We adjust the pre-selected size of the training dataset composed of Doc-Param pairs, increasing it from 480 to 4800. As shown in Figure~\ref{fig:llama_vary} and~\ref{fig:qwen_vary}, DyPRAG achieves strong performance even with just 480 training examples. The performance remains remarkably stable across different dataset sizes, indicating that our design, $\mathcal{F}^\prime_\phi$, is capable of learning the underlying mapping between documents and parameters with minimal data.

% \begin{table}[t]
% \small
% \centering
% \caption{Supplement ablation study of alignment loss. The backbone model is the LLaMA3.2-1B.}
% \label{tab:dyprag-performance-sup}
% \resizebox{0.6\textwidth}{!}{
% \begin{tabular}{@{}lccccc@{}}
% \toprule
% \multirow{2}{*}{\textbf{Methods}} & \multicolumn{5}{c}{\textbf{2WQA}}  \\
% \cmidrule(lr){2-6}
%  & \textbf{Compare} & \textbf{Bridge} & \textbf{Inference} & \textbf{Compose} & \textbf{Total} \\
% \midrule
% \textbf{DyPRAG} & 0.5125 & \textbf{0.4815} & \textbf{0.1735} & 0.0754 & 0.2531 \\
% \rowcolor{gray!20} \multicolumn{6}{@{}l}{\textit{\textbf{Ablation Study}}} \\
% w/o $\mathcal{L}_{kl}$ & 0.2954 & 0.3774 & 0.1294 & 0.0578 & 0.2027 \\
% w/o $\mathcal{L}_{mse}$ & \textbf{0.5606} & 0.3696 & 0.1729 & \textbf{0.084} & \textbf{0.2728}     \\
% w/o $\mathcal{L}_{kl},\mathcal{L}_{mse}$ & 0.4523 & 0.2484 & 0.1674 & 0.0748 & 0.2343 \\
% \bottomrule
% \end{tabular}}
% \vspace{-0.2cm}
% \end{table}

\begin{table}[t]

\caption{Ablation study of alignment loss. The backbone model is the LLaMA3.2-1B.}
\label{tab:llama_loss}
\centering
\resizebox{\textwidth}{!}{
\begin{tabular}{@{}lc*{5}c*{3}ccc@{}}
    \toprule
    \multirow{2}{*}{\textbf{Methods}} & \multicolumn{5}{c}{\textbf{2WQA}} & \multicolumn{3}{c}{\textbf{HQA}} & \multirow{2}{*}{\textbf{PQA}} & \multirow{2}{*}{\textbf{CWQ}}  & \multirow{2}{*}{\textbf{Avg}}
    \\
    \cmidrule(lr){2-6} \cmidrule(lr){7-9} 
    &  
    \textbf{Compare} & \textbf{Bridge} & \textbf{Inference} & \textbf{Compose} & \textbf{Total}   & 
    \textbf{Bridge} & \textbf{Compare} & \textbf{Total}  & & &  \\
    \midrule
    \textbf{DyPRAG} & 51.25 & \textbf{48.15} & \textbf{17.35} & 7.54 & 25.31 & \textbf{14.05} & \textbf{43.9} & \textbf{19.97} & \textbf{8.37} & \textbf{36.86} & \textbf{25.28}\\ 
    \rowcolor{gray!20} \multicolumn{12}{@{}l}{\textit{\textbf{Ablation Study}}} \\ 
    w/o $\mathcal{L}_{kl}$ &  29.54 & 37.74 & 12.94 & 5.78 & 20.27 & 7.35 & 35.28 & 13.12 & 1.93 & 22.85 & 18.68 \\ 
    w/o $\mathcal{L}_{mse}$ &  \textbf{56.06} & 36.96 & 17.29 & \textbf{8.40} & \textbf{27.28} & 12.78 & 42.11 & 17.65 & 5.94 & 32.98 & 23.38\\ 
    w/o $\mathcal{L}_{kl},\mathcal{L}_{mse}$ &  45.23 & 24.84 & 16.74 & 7.48 & 23.43 &  12.66 & 39.46 & 18.26 & 2.42 & 34.92 & 22.54 \\ 
    \bottomrule
\end{tabular}
}
\end{table}

\paragraph{\noindent\textbf{Effect of Alignment Loss.}}
In Section~\ref{sec:train}, our alignment loss $\mathcal{L}_\text{align}$ is composed of three components: $\mathcal{L}_\text{pred}$, $\mathcal{L}_\text{mse}$, and $\mathcal{L}_\text{kl}$. To investigate which component contributes most to the effectiveness of DyPRAG, we conducted ablation studies. As shown in Table~\ref{tab:llama_loss}, removing any single loss component negatively impacts the model's performance. For instance, when $\mathcal{L}_\text{kl}$ is removed, the model's performance drops significantly, demonstrating that aligning with the target model's output distribution is an effective strategy~\cite{liao2025instancetraininginstructionlearning}. While removing $\mathcal{L}_\text{mse}$ has the smallest impact, ensuring that $\mathcal{F}^\prime_{\phi}$ generates $p^\prime$ values as close as possible to the trained $p$ still proves beneficial for DyPRAG. Furthermore, even when only $\mathcal{L}_\text{pred}$ is retained, DyPRAG maintains stable performance, indicating that the standard language modeling loss plays a central role in the overall alignment process.

\paragraph{\noindent\textbf{Comparison of Response Length.}}

\begin{figure}[t]
    \begin{minipage}{0.53\textwidth}
         \centerline{\includegraphics[width=1\textwidth]{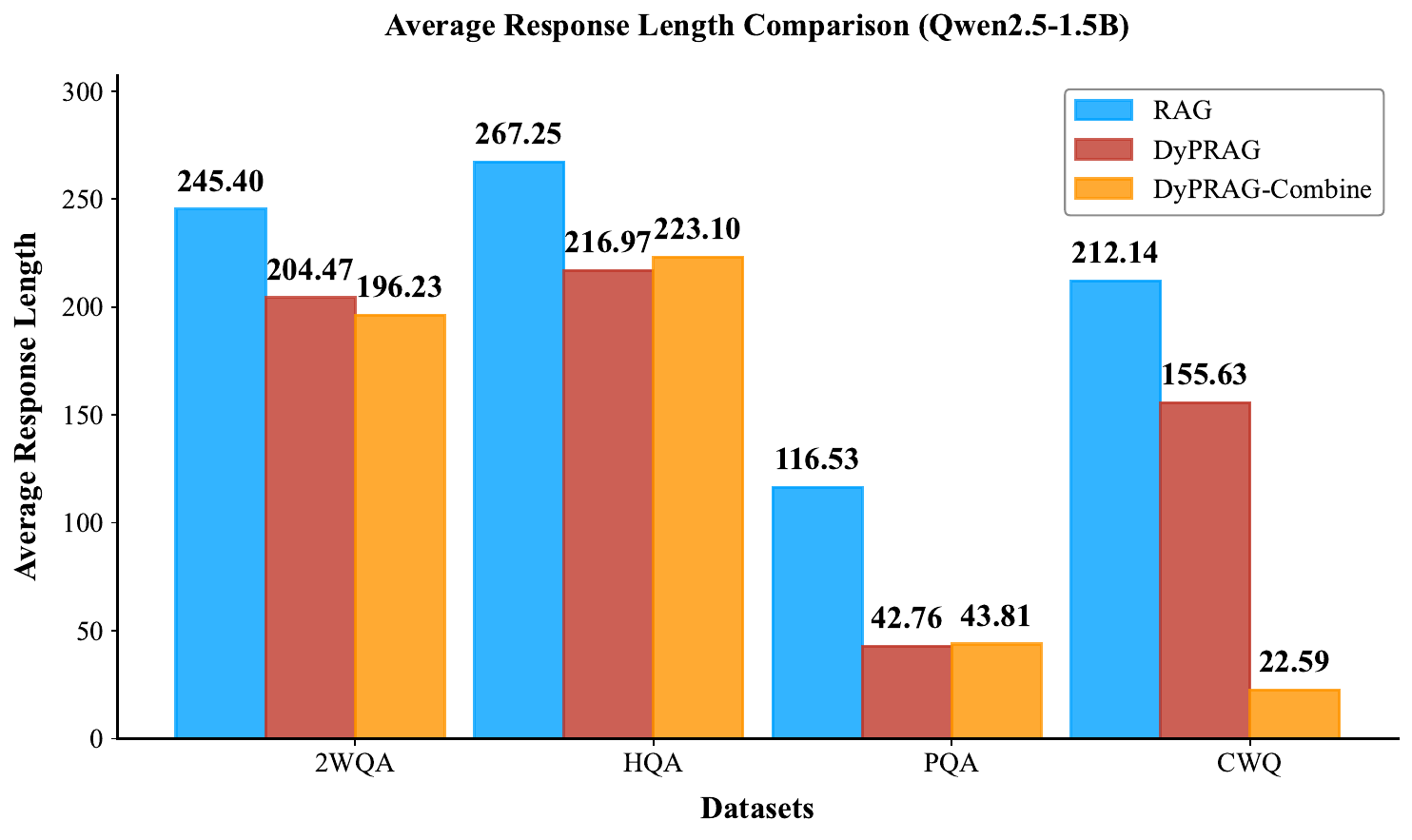}}
        \caption{
        Comparison of response length across various datasets. The backbone model is the Qwen2.5-1.5B. 
        }
        \label{fig:qwen_res_length}
    \end{minipage}
    \hfill
    \begin{minipage}{0.45\textwidth}
        \centerline{\includegraphics[width=0.9\textwidth]{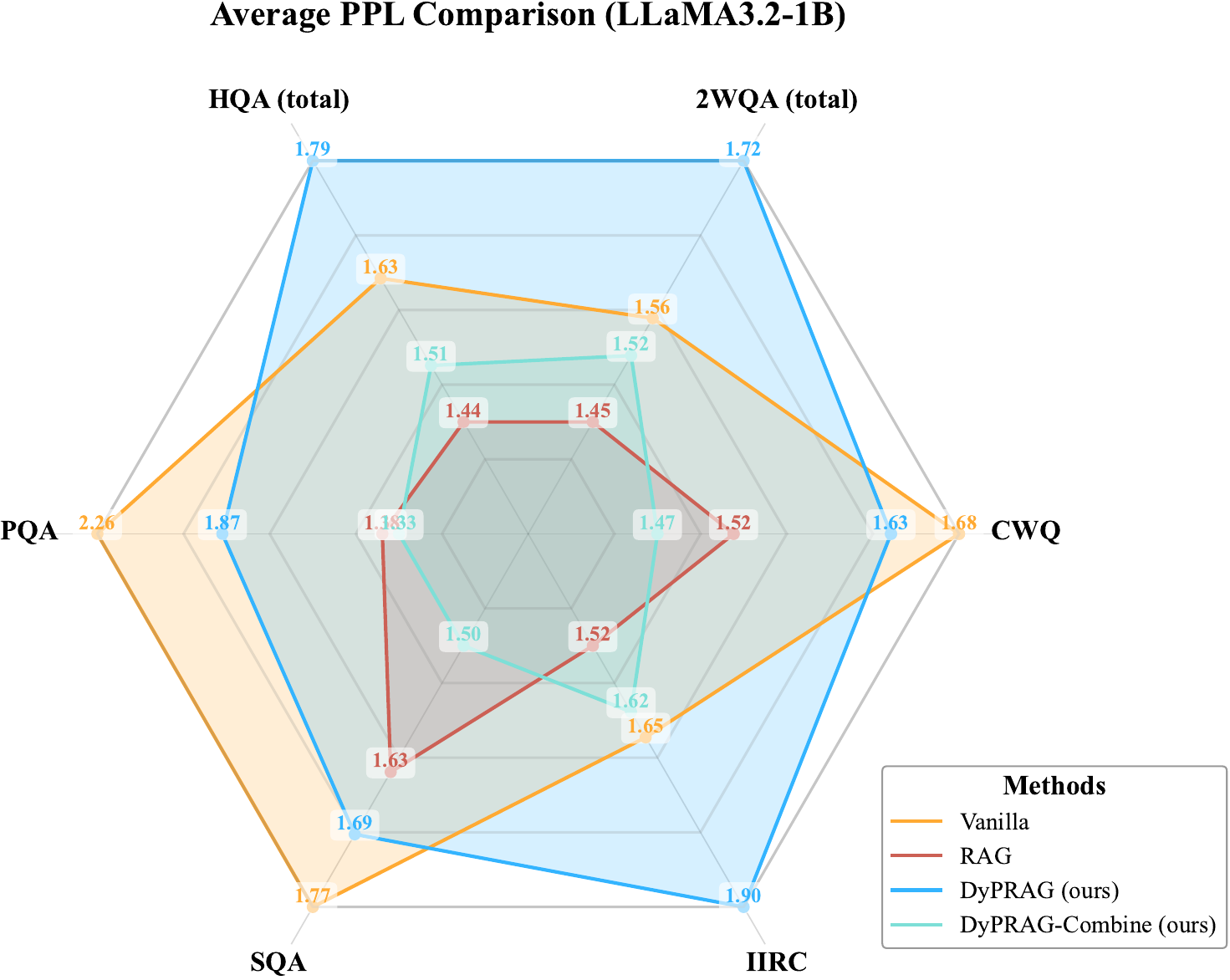}}
        \caption{
        Comparison of average PPL. Smaller PPL means less conflict. The backbone model is the LLaMA3.2-1B. 
        }
        \label{fig:ppl}

    \end{minipage}
\end{figure}

Notably, we consider only the context length when calculating inference cost. However, in practice, the response length from LLMs also affects inference time. As shown in Figure~\ref{fig:qwen_res_length}, we compare DyPRAG-Combine with RAG across four benchmarks, considering the average response length. DyPRAG-Combine significantly reduces response length, by 20\% in 2WQA and up to 90\% in CWQ. This demonstrates that DyPRAG-Combine can answer questions correctly with fewer tokens, thereby lowering inference costs and avoiding redundant information.

\section{Exploring Metrics for RAG Hallucination Detection}
\label{app:rag_hal}
\paragraph{\noindent\textbf{Can Perplexity Reflects Knowledge Conflict?}} Recent studies have explored methods to detect hallucinations in LLMs and RAG systems by leveraging various metrics~\cite{chen2024inside,sun2025redeepdetectinghallucinationretrievalaugmented}. Among these, we first adopt the simplest yet effective metric which only need single generation, Perplexity (PPL)~\cite{ren2022out}, to evaluate knowledge conflicts. As illustrated in Figure\ref{fig:ppl}, Vanilla and DyPRAG exhibit higher PPL, while DyPRAG-Combine and RAG demonstrate significantly lower PPL. However, these results are inconsistent with the findings in Table~\ref{tab:main} and Table~\ref{tab:ood_performance}. For instance, although DyPRAG-Combine achieves the best performance on IIRC, its calculated PPL suggests a higher probability of knowledge conflicts, which is clearly incorrect. We hypothesize that this discrepancy primarily stems from variations in model parameters introduced by parameter injection in DyPRAG, which cannot be detected using the simple PPL method. Given that different tokens contribute unequally to the overall semantics of a sentence, the PPL, which calculated as the average of token-level uncertainty, fails to effectively capture the uncertainty of the entire sequence.

\paragraph{\noindent\textbf{Effective Detection with Sentence-Level Metrics.}}
Given the limitations of PPL, we decided to explore alternative metrics that leverage multiple generations. Research has shown that generating multiple outputs for a single input is beneficial for estimating sequence-level uncertainty. To this end, we set the temperature to 1.0, top\_p to 0.95, and top\_k to 20, generating five responses to calculate Entropy (EN), Length Normalized Entropy (LEN)~\cite{malinin2020uncertainty}, and Lexical Similarity (LS)~\cite{lin2022towards} to evaluate the probability of RAG hallucination\footnote{We use the implementation in \url{https://github.com/alibaba/eigenscore}}. As shown in Table~\ref{tab:multigen}, our approach demonstrates reduced knowledge conflict in most scenarios, especially in our strongest DyPRAG-Combine.

We observed that both EN and LEN increase when in-context injection is applied, suggesting that in RAG systems, the retrieved passages often conflict with the model's internal knowledge. In contrast, utilizing DyPRAG to inject converted parametric knowledge significantly reduces the likelihood of knowledge conflict, demonstrating the effectiveness of DyPRAG. However, the LS results indicate that adding context reduces conflict, which contradicts the established definition of RAG hallucination. We argue that EN and LEN are more suitable for effective RAG hallucination detection in DyPRAG settings. Exploring more effective detection methods remains an important direction for future work.

% \begin{wrapfigure}{r}{0.4\textwidth}

% \end{wrapfigure}

\begin{table*}[t]
\caption{We present the experimental results for the knowledge conflict metrics of DyPRAG and DyPRAG-Combine, in comparison with Vanilla and Standard RAG. Three multi-generation metrics are employed: \textbf{Entropy (EN)}, \textbf{Length Normalized Entropy (LEN)}, and \textbf{ Lexical Similarity (LS)}. In these metrics, $\uparrow$ indicates that higher values are better, while $\downarrow$ indicates the opposite. The best performance for each metric is highlighted in bold. The backbone model is the LLaMA3.2-1B.}
\label{tab:multigen}
\small
\centering

\setlength{\extrarowheight}{2pt} % 增加表格行距
\resizebox{\textwidth}{!}{
\begin{tabular}{@{}lcc{c}ccccc@{}}
\toprule
\textbf{Metric} & 
\makecell[c]{\textbf{Method}} & 
\textbf{2WQA (total)} & 
\textbf{HQA (total)} & 
\textbf{PQA} & 
\textbf{CWQ} &
\textbf{SQA} &
\textbf{IIRC}
\\
\midrule
\multirow{4}{*}{\textbf{EN} $\downarrow$} & 
\textbf{Vanilla} & 
~3.187 & ~3.176 & ~3.251 & ~3.163 & ~3.178 & ~3.011 \\
& \textbf{DyPRAG (ours)} & 
~\textbf{2.199} & ~\textbf{1.999} & ~\textbf{1.757} & ~\textbf{2.860} & ~\textbf{2.805} & ~\textbf{2.544}  \\

\cdashline{3-8} 

& \textbf{Standard RAG} & 
~3.565 & ~3.453 & ~3.778 & ~3.619 & ~3.398 & ~3.030 \\
& \textbf{DyPRAG-Combine (ours)} & 
~\textbf{2.755} & ~\textbf{2.470} & ~\textbf{3.584} & ~\textbf{3.467} & ~\textbf{3.136} & ~\textbf{2.555} \\

\midrule

\multirow{4}{*}{\textbf{LEN} $\downarrow$} & 
\textbf{Vanilla} & 
~0.637 & ~0.635 & ~0.650 & ~0.633 & ~0.636 & ~0.602 \\
& \textbf{DyPRAG (ours)} & 
~\textbf{0.440} & ~\textbf{0.400} & ~\textbf{0.586} & ~\textbf{0.572} & ~\textbf{0.561} & ~\textbf{0.509}  \\

\cdashline{3-8}

& \textbf{Standard RAG} & 
~0.713 & ~0.691 & ~0.756 & ~0.724 & ~0.680 & ~0.606 \\
& \textbf{DyPRAG-Combine (ours)} & 
~\textbf{0.551} & ~\textbf{0.494} & ~\textbf{0.719} & ~\textbf{0.693} & \textbf{0.627} & \textbf{0.511} \\

\midrule

\multirow{4}{*}{\textbf{LS} $\uparrow$} & 
\textbf{Vanilla} & 
~\textbf{0.923} & ~\textbf{0.936} & ~0.723 & ~0.730 &  ~0.497 & ~0.963 \\
& \textbf{DyPRAG (ours)} & 
~0.915 & ~0.933 & ~\textbf{0.842}  & ~\textbf{0.859} & ~\textbf{0.527} & ~\textbf{0.966}  \\
\cdashline{3-8}
&\textbf{Standard RAG} & 
~0.945 & ~0.956 & ~0.936 & ~0.962 & ~0.812 & ~0.966 \\
& \textbf{DyPRAG-Combine (ours)} & 
~\textbf{0.953} & ~\textbf{0.959} & ~\textbf{0.966} & ~\textbf{0.988} & ~\textbf{0.853} & ~\textbf{0.975} \\

\bottomrule
\end{tabular}%
}
\end{table*}

\section{Why Vanilla Outperforms RAG Occasionally ?}
\label{app:rag_deficiency}
In this section, we provide a detailed analysis of why the vanilla model occasionally outperforms RAG. As shown in Table~\ref{tab:main}, the vanilla model surpasses RAG most significantly in 2WQA, although the results vary across different models. For instance, the vanilla model outperforms RAG by 2.62 and 0.99 in Qwen2.5-1.5B and LLaMA3-8B on average, respectively. After analyzing the cases, we identify two key issues that affect RAG's performance: \textbf{1) Poor Retriever.} Following~\cite{su2025parametricretrievalaugmentedgeneration}, we use BM25 as the retriever. However, in many cases, the retrieved documents contain only similar words rather than relevant content. This results in the provided content being unhelpful or even detrimental to the performance of LLMs. \textbf{2) Already Seen Data.} During the pre-training stages of the selected LLMs~\cite{qwen2025qwen25technicalreport,Llama-3-8B-Instruct,Llama-3.2-1B-Instruct}, the external source we use (i.e., Wikipedia) has already been seen. This allows LLMs to answer certain questions independently, especially in simpler tasks like 2WQA. Moreover, the inclusion of incorrect or irrelevant context further degrades the performance, as observed in Table~\ref{tab:main}.

A more rigorous evaluation setting should include ground-truth passages and ensure no or less data leakage.Under this setting, as shown in Table~\ref{tab:ood_performance}, the performance of the vanilla model is significantly lower than that of RAG, which aligns with our hypothesis. For instance, the vanilla model achieves only 8.78\% and 1.00\% accuracy on Qwen2.5-1.5B for IIRC and SQA, respectively. In contrast, DyPRAG demonstrates a notable improvement in test-time knowledge, achieving 10.23\% and 15.67\% accuracy on Qwen2.5-1.5B for IIRC and SQA, respectively. These results underscore the critical role of RAG while showcasing the ability of our proposed DyPRAG to seamlessly enhance OOD knowledge effectively. Furthermore, DyPRAG-Combine establishes a superior RAG paradigm by delivering even better performance under these more challenging conditions.
In summary, we believe that this more rigorous experimental setting better validates our proposed method.

\section{Further Analysis of Contextual and Parametric Knowledge Conflict }
\label{app:context_param_ana}

\paragraph{\noindent\textbf{Parameter Injection Makes LLMs Trust Themselves.}}
As shown in Table~\ref{tab:conflict_case_2}, while vanilla LLMs contain accurate parametric knowledge regarding which director was born later, the introduction of retrieved documents about each director causes contextual knowledge to mislead $\mathcal{M}$, resulting in the incorrect answer "William Lustig" while DyPRAG stays the same.  This demonstrates that DyPRAG can effectively reduce the knowledge conflict problem. In this case, standard RAG often introduces redundant or incorrect information from the context, a phenomenon commonly referred to as hallucination~\cite{sun2025redeepdetectinghallucinationretrievalaugmented}. In contrast, our proposed DyPRAG effectively incorporates accurate information into parametric knowledge. This allows DyPRAG-Combine to align parametric knowledge with contextual knowledge, thereby reducing the likelihood of hallucinations and enabling LLMs to rely more consistently on its own knowledge.

\paragraph{\noindent\textbf{Dynamic Parametric Knowledge Enhances LLMs at Test-time.}} Our DyPRAG serves as an effective plug-and-play technique for enhancing knowledge during test-time. As demonstrated in Table~\ref{tab:conflict_case_3}, DyPRAG successfully manipulates the original parametric knowledge of large language models (LLMs) in 14.67\% of cases. Therefore, it can directly enhance the model's knowledge during inference without the need for further fine-tuning.

\paragraph{\noindent\textbf{Proportion of Different Combinations.}}
Furthermore, as shown in Table~\ref{tab:confilct_rate}, when both Vanilla LLMs and RAG give incorrect answers, DyPRAG provides the correct answer 26.33\% of the time. This indicates that DyPRAG can effectively inject missing parametric knowledge and outperforms in-context injection methods. Additionally, in cases where the vanilla LLM provides the correct answer (i.e., the model possesses accurate internal knowledge), RAG achieves a correct answer rate of 5.33\%, while DyPRAG performs better with a rate of 6.33\%, showing that parameter injection leads to lower conflict. Similar trend of DyPRAG-Combine is presented  in Table~\ref{tab:confilct_rate_combine}.

These results demonstrate that our proposed DyPRAG injects parametric knowledge successfully and mitigates conflicts between internal parametric knowledge and external contextual knowledge through the injection of knowledgeable LoRA adapters.

\begin{figure}[htbp]
    \centering
\begin{minipage}{0.45\textwidth}
    \centering
    \caption{Case study about contextual and parametric knowledge conflict in 2WQA (Bridge sub-task) where only standard RAG answers wrongly (6.67\%). The backbone model is the LLaMA3.2-1B. 
% \textbf{Vanilla} represents the answer from original LLMs without any external knowledge.
\colorbox{red!30}{\textcolor{red!30}{1}}:deficiency in parametric knowledge,
\colorbox{yellow!60}{\textcolor{yellow!60}{1}}: knowledge conflict, \colorbox{green!30}{\textcolor{green!30}{1}}: successful knowledge manipulation.
}
\label{tab:conflict_case_2}
    \resizebox{\textwidth}{!}{%
\begin{tabular}{lcc}
\hline
\multicolumn{3}{p{0.95\linewidth}}{\textbf{Question:} Which film has the director born later, \colorbox{yellow!60}{Diary Of A Maniac} or \colorbox{green!30}{Return Of The Hero}?} \\
\hline
\multicolumn{3}{p{0.95\linewidth}}{\textbf{Ground truth:} \colorbox{green!30}{Return Of The Hero}} \\
\hline
\multicolumn{3}{p{0.95\linewidth}}{\textbf{Retrieved top-1 document:} Maniac (1980 film) Maniac is a 1980 American psychological slasher film directed by \colorbox{yellow!60}{William Lustig} and written by C. A. Rosenberg...} \\
\hline
\textbf{Method} & \textbf{Answer} & \textbf{Status} \\
\hline
\textbf{Vanilla} & \colorbox{green!30}{Return Of The Hero }&  \textcolor{green}{\Checkmark}\\
\hline
\textbf{Standard RAG} & \colorbox{yellow!60}{William Lustig} & \textcolor{red}{\textsf{\XSolidBrush}} \\
\hline
\textbf{DyPRAG} (ours) & \colorbox{green!30}{Return Of The Hero} &  \textcolor{green}{\Checkmark} \\
\textbf{DyPRAG-Combine} (ours)  & \colorbox{green!30}{Return Of The Hero} & \textcolor{green}{\Checkmark} \\
\hline
\end{tabular}
}
\end{minipage}
\hfill
\begin{minipage}{0.48\textwidth}
    \centering
    
\caption{Case study about contextual and parametric knowledge conflict in 2WQA (Bridge sub-task) where only DyPRAG and DyPRAG-Combine answer wrongly (14.67\%). The backbone model is the LLaMA3.2-1B. 
% \textbf{Vanilla} represents the answer from original LLMs without any external knowledge.
\colorbox{red!30}{\textcolor{red!30}{1}}:deficiency in parametric knowledge,
\colorbox{yellow!60}{\textcolor{yellow!60}{1}}: knowledge conflict, \colorbox{green!30}{\textcolor{green!30}{1}}: successful knowledge manipulation
}
\label{tab:conflict_case_3}
    \resizebox{\textwidth}{!}{%
\begin{tabular}{lcc}
\hline
\multicolumn{3}{p{0.95\linewidth}}{\textbf{Question:} Which film has the director born later, \colorbox{yellow!60}{Miss Sloane} or \colorbox{green!30}{Time Changer}?} \\
\hline
\multicolumn{3}{p{0.95\linewidth}}{\textbf{Ground truth:} \colorbox{green!30}{Time Changer}} \\
\hline
\multicolumn{3}{p{0.95\linewidth}}{\textbf{Retrieved top-1 document:} production budget of \$13 million. "\colorbox{yellow!60}{Miss Sloane}" is ranked number 75 by per-theater average on Box Office...} \\
\hline
\textbf{Method} & \textbf{Answer} & \textbf{Status} \\
\hline
\textbf{Vanilla} & \colorbox{red!30}{John Frankenheimer}&  \textcolor{red}{\textsf{\XSolidBrush}}\\
\hline
\textbf{Standard RAG} & \colorbox{yellow!60}{Miss Sloane} & \textcolor{red}{\textsf{\XSolidBrush}} \\
\hline
\textbf{DyPRAG} (ours) & \colorbox{green!30}{Time Changer} &  \textcolor{green}{\Checkmark} \\
\textbf{DyPRAG-Combine} (ours)  & \colorbox{green!30}{Time Changer} & \textcolor{green}{\Checkmark} \\
\hline
\end{tabular}
}
        
\end{minipage}
\end{figure}

\begin{figure}[hbp]
    \begin{minipage}{0.48\textwidth}
    \centering
    \caption{Right/Wrong answer combinations of Vanilla, RAG, DyPRAG and corresponding proportional distribution in 2WQA (Bridge Sub-task). The backbone model is the LLaMA3.2-1B. \textcolor{green}{\textsf{\Checkmark}} indicates a correct answer, while \textcolor{red}{\textsf{\XSolidBrush}} indicates an incorrect answer. The "Ratio (\%)" column on the right represents the percentage of each combination across the dataset (300 examples).}
\label{tab:confilct_rate}
        \begin{tabular}{@{}cccc@{}}
\toprule
Vanilla & RAG & DyPRAG & Ratio(\%)  \\
\hline
\textcolor{green}{\Checkmark} & \textcolor{green}{\Checkmark} & \textcolor{green}{\Checkmark} & 4.67 \\

\textcolor{red}{\textsf{\XSolidBrush}}  & \textcolor{red}{\textsf{\XSolidBrush}}  & \textcolor{red}{\textsf{\XSolidBrush}}  & 34.67 \\

\textcolor{green}{\Checkmark}& \textcolor{red}{\textsf{\XSolidBrush}} & \textcolor{green}{\Checkmark} & \textbf{6.33}\\

\textcolor{green}{\Checkmark}& \textcolor{green}{\Checkmark} & \textcolor{red}{\textsf{\XSolidBrush}} & 5.33\\

\textcolor{red}{\textsf{\XSolidBrush}} & \textcolor{green}{\Checkmark} & \textcolor{green}{\Checkmark} & 8.33\\

\textcolor{red}{\textsf{\XSolidBrush}} & \textcolor{red}{\textsf{\XSolidBrush}} & \textcolor{green}{\Checkmark} & \textbf{26.33}\\

\textcolor{red}{\textsf{\XSolidBrush}} & \textcolor{green}{\Checkmark} & \textcolor{red}{\textsf{\XSolidBrush}}  & 7.67\\

\textcolor{green}{\Checkmark} & \textcolor{red}{\textsf{\XSolidBrush}} & \textcolor{red}{\textsf{\XSolidBrush}} & 6.33\\
\bottomrule
\end{tabular}

    \end{minipage}
    \hfill
    \begin{minipage}{0.48\textwidth}
    \caption{Right/Wrong answer combinations  of Vanilla, RAG, DyPRAG-Combine and corresponding proportional distribution in 2WQA (Bridge Sub-task). The backbone model is the LLaMA3.2-1B. \textcolor{green}{\textsf{\Checkmark}} indicates a correct answer, while \textcolor{red}{\textsf{\XSolidBrush}} indicates an incorrect answer. The "Ratio (\%)" column on the right represents the percentage of each combination across the dataset (300 examples).}
    \label{tab:confilct_rate_combine}
    \centering
    \begin{tabular}{@{}cccc@{}}
    \toprule
    Vanilla & RAG & DyPRAG-Combine & Ratio(\%)  \\
    \hline
    \textcolor{green}{\Checkmark} & \textcolor{green}{\Checkmark} & \textcolor{green}{\Checkmark} & 5.33 \\
    
    \textcolor{red}{\textsf{\XSolidBrush}}  & \textcolor{red}{\textsf{\XSolidBrush}}  & \textcolor{red}{\textsf{\XSolidBrush}}  & 35.00 \\
    
    \textcolor{green}{\Checkmark}& \textcolor{red}{\textsf{\XSolidBrush}} & \textcolor{green}{\Checkmark} & \textbf{6.33}\\
    
    \textcolor{green}{\Checkmark}& \textcolor{green}{\Checkmark} & \textcolor{red}{\textsf{\XSolidBrush}} & 4.67\\

    \textcolor{red}{\textsf{\XSolidBrush}} & \textcolor{green}{\Checkmark} & \textcolor{green}{\Checkmark} & 8.00\\
    
    \textcolor{red}{\textsf{\XSolidBrush}} & \textcolor{red}{\textsf{\XSolidBrush}} & \textcolor{green}{\Checkmark} & \textbf{26.00}\\
    
    \textcolor{red}{\textsf{\XSolidBrush}} & \textcolor{green}{\Checkmark} & \textcolor{red}{\textsf{\XSolidBrush}}  & 8.00\\
    
    \textcolor{green}{\Checkmark} & \textcolor{red}{\textsf{\XSolidBrush}} & \textcolor{red}{\textsf{\XSolidBrush}} & 6.67\\
    \bottomrule
\end{tabular}

    \end{minipage}
\end{figure}

\clearpage  

\section{Visualization of Parameter Translator Workflow.}
\label{app:visual}
\begin{figure*}[htbp]
\centerline{\includegraphics[width=\textwidth]{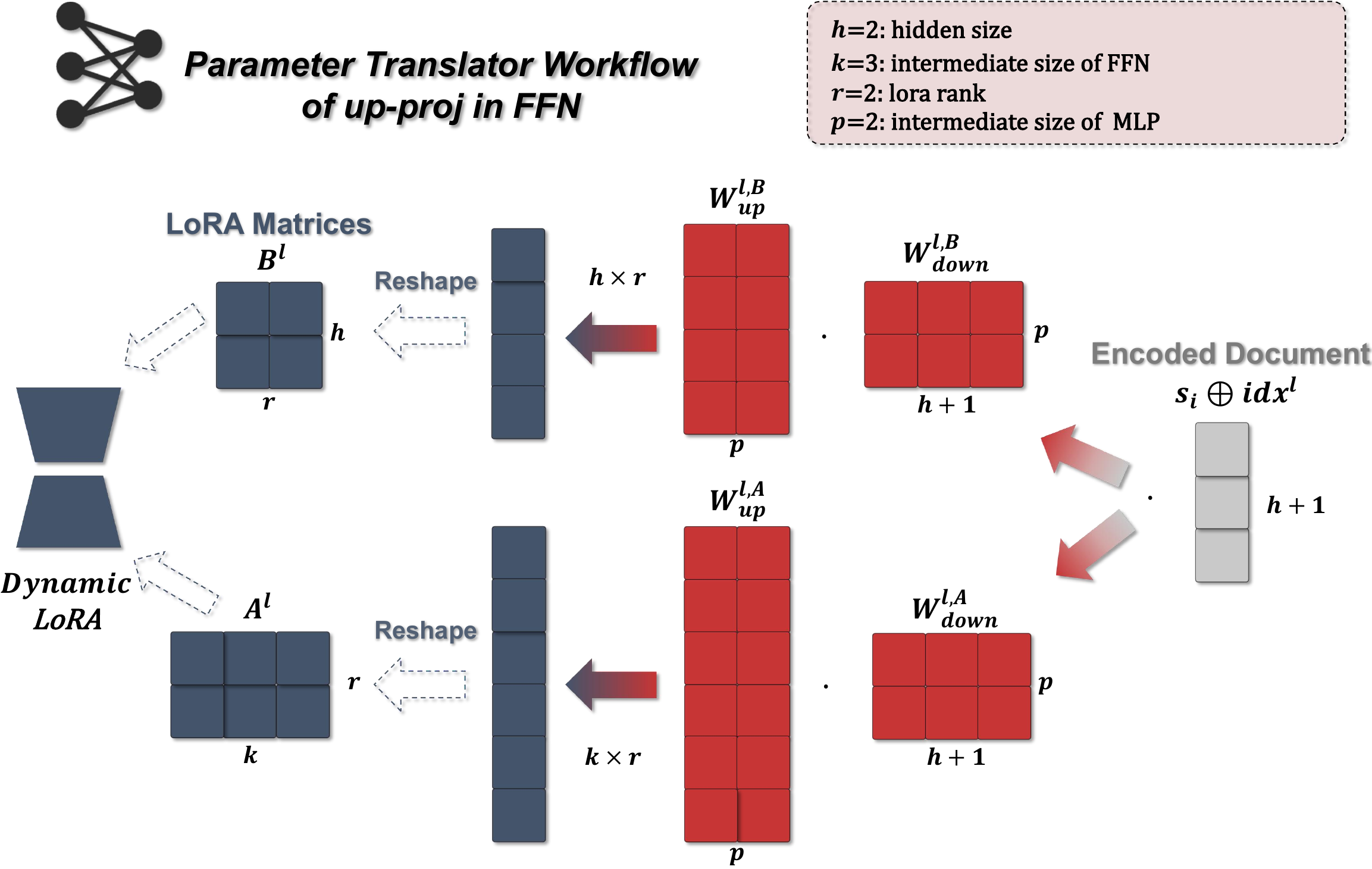}}
% \vspace{-0.45cm}
\caption{
Visualization of the parameter translator workflow of up-project in FFN. The overall process remains consistent with Eq.~\ref{eq:parameter_translator}.
}
\label{fig:parameter_translator}
\end{figure*}

To clearly illustrate the workflow of the parameter translator $\mathcal{F}^\prime_\phi$, we use the up-proj module in the FFN as an example, as shown in Figure~\ref{fig:parameter_translator}. This visualization demonstrates the transformation of document embeddings into dynamic LoRAs, consistent with Eq.~\ref{eq:parameter_translator}.

\newpage

\section{Prompt for Main Experiments Evaluation}
In the main experiments, we used the following prompt to assess the performance of DyPRAG and other baseline models in Figure~\ref{fig:nocot_format} and ~\ref{fig:cot_format}:

\begin{figure*}[htbp]
\begin{tcolorbox}[colback=lightgray!20,colframe=darkgray!80,title=Prompt Format of No-CoT]
\small
You should answer the question by referring to the knowledge provided below and integrating your own knowledge.
\\
\\
Passage 1: \{passages[0]\} \\
Passage 2: \{passages[1]\} \\
Passage 3: \{passages[2]\} 
\\
\\
Question: \{question\}

The answer is \{answer\}
\end{tcolorbox}
\caption{Prompt format of No-CoT in our expriments.}
\label{fig:nocot_format}
\end{figure*}

\begin{figure*}[htbp]
\begin{tcolorbox}[colback=lightgray!20,colframe=darkgray!80,title=Prompt Format of CoT]
\small
You should reference the knowledge provided below and combine it with your own knowledge to answer the question. Please follow the format of the example I provided above.
Here are some examples about how to answer the questions.

Question: $\text{fewshot}_{q}[0]$

Answer: $\text{fewshot}_{a}[0]$

Question: $\text{fewshot}_{q}[1]$ 

Answer: $\text{fewshot}_{a}[1]$ 

Question: $\text{fewshot}_{q}[2]$ 

Answer: $\text{fewshot}_{a}[2]$ 

...\\

Here are some reference.

Passage 1: \{passages[0]\} \\
Passage 2: \{passages[1]\} \\
Passage 3: \{passages[2]\} 
\\
\\
Let's think step by step. Answer the questions in the same format as above.

Question: \{question\}

Answer: \{answer\}
\end{tcolorbox}
\caption{Prompt format of CoT in our expriments.}
\label{fig:cot_format}
\end{figure*}
\newpage

\section{Prompt for Knowledge Internalization Evaluation}
In the knowledge internalization experiments, we used the following prompt to assess the internalization ability of RAG generation from DyPRAG-Combine and standard RAG method evaluated by GPT-4o in Figure~\ref{fig:ragtruth_format}:

\begin{figure*}[htbp]
\begin{tcolorbox}[colback=lightgray!20,colframe=darkgray!80,title=Prompt Format of Evaluate RAGTruth]
\small
Compare DyPRAG and RAG answers to assess which better internalizes knowledge—integrating its own knowledge with the given context for a natural, informed response. \\

Evaluation Criteria: \\
1. Internalization: Does the answer go beyond repetition to integrate knowledge seamlessly? \\
2. Fluency: Is the response well-structured and readable? \\
3. Relevance: Does it stay on topic while demonstrating depth? \\

Mark the Winner: Identify the superior response. If both are equally strong, mark it as a tie. \\

Question: \{question\}

Context: \{passages\}

DyPRAG Answer: \{dyprag\_answer\}

RAG Answer: \{rag\_answer\} \\

Respond in the following format: \\ 
\{\{ \\ 
  "win model": "DyPRAG or RAG or Tie", \\
  "reason": "Provide a concise explanation of why the selected answer demonstrates better knowledge integration, referencing the question, context, and specific details from both answers. If one answer has clear advantages in integration, explain them; if there are errors or weaknesses, specify them." \\
\}\}
\end{tcolorbox}
\caption{Prompt format of evaluate RAGTruth using GPT-4o. We compare answer between standard RAG and DyPRAG-Combine.}
\label{fig:ragtruth_format}
\end{figure*}

\begin{figure*}[htbp]
\begin{tcolorbox}[colback=lightgray!20,colframe=darkgray!80,title=Prompt Format of QA Generation]
\small

I will provide a passage of text, and you need to generate three different questions based on the content of this passage. Each question should be answerable using the information provided in the passage. Additionally, please provide an appropriate answer for each question derived from the passage.

You need to generate the question and answer in the following format:

[

    \{\{
    
        "question": "What is the capital of France?",
        
        "answer": "Paris",
        
        "full\_answer": "The capital of France is Paris."
    
    \}\}, 

]

This list should have at least three elements. You only need to output this list in the above format.
Passage:
\{passage\}
\end{tcolorbox}
\caption{Prompt format of evaluate RAGTruth using GPT-4o. We compare answer between standard RAG and DyPRAG-Combine.}
\label{fig:ragtruth_format}
\end{figure*}

\section{Limitations and Future Directions}
\label{limitation}
In this study, our proposed Dynamic Parametric RAG (DyPRAG) demonstrates superior performance in both IID and OOD settings across various scales of LLMs. However, it is important to note that our method focuses exclusively on the QA task. For other challenging tasks, such as mathematical reasoning, this remains a highly valuable area for further exploration.
Additionally, as we propose combining DyPRAG generated parameters with contextual knowledge to mitigate the RAG hallucination issue, it becomes crucial to conduct in-depth analysis using interpretability tools to better understand the internal workflow of the model.
Finally, our research represents only the beginning of exploring the potential for a powerful and trustworthy RAG solution. Such a system could seamlessly enhance parametric knowledge with contextual information, improving knowledge fusion while mitigating conflicts. Developing and deploying this RAG system in real-world applications is a promising and worthwhile avenue for future work.

\end{document}